\documentclass{article}

\usepackage{hyperref}
\usepackage{url}
\usepackage{graphicx}
\usepackage{subcaption}
\usepackage{multirow} 
\usepackage{float}
\usepackage{booktabs}
\usepackage{placeins}
\usepackage{needspace}
\usepackage{fvextra}
\usepackage{float}

\usepackage[accepted]{icml2026}

\usepackage{amsmath}
\usepackage{amssymb}
\usepackage{mathtools}
\usepackage{amsthm}

\usepackage[capitalize,noabbrev]{cleveref}

\theoremstyle{plain}

\theoremstyle{definition}

\theoremstyle{remark}

\usepackage[disable,textsize=tiny]{todonotes}
\icmltitlerunning{Uncovering Grounding IDs: How External Cues Shape Multi-Modal Binding}

\begin{document}

\twocolumn[
  \icmltitle{Uncovering Grounding IDs: How External Cues Shape Multi-Modal Binding}

  % It is OKAY to include author information, even for blind submissions: the
  % style file will automatically remove it for you unless you've provided
  % the [accepted] option to the icml2026 package.

  % List of affiliations: The first argument should be a (short) identifier you
  % will use later to specify author affiliations Academic affiliations
  % should list Department, University, City, Region, Country Industry
  % affiliations should list Company, City, Region, Country

  % You can specify symbols, otherwise they are numbered in order. Ideally, you
  % should not use this facility. Affiliations will be numbered in order of
  % appearance and this is the preferred way.
  \icmlsetsymbol{equal}{*}

\begin{icmlauthorlist}
  \icmlauthor{Amirmohammad Izadi}{equal,sharif}
  \icmlauthor{Hosein Hasani}{equal,sharif}
  \icmlauthor{Fatemeh Askari}{equal,sharif}
  \icmlauthor{Mobin Bagherian}{equal,sharif}
  \icmlauthor{Sadegh Mohammadian}{sharif}
  \icmlauthor{Mohammad Izadi}{sharif}
  \icmlauthor{Mahdieh Soleymani Baghshah}{sharif}
\end{icmlauthorlist}

\icmlaffiliation{sharif}{Department of Computer Engineering, Sharif University of Technology, Tehran, Iran}

% \icmlEqualContribution{Amirmohammad Izadi and Hosein Hasani contributed equally as first authors.}
% \icmlEqualContribution{Fatemeh Askari and Mobin Bagherian contributed equally as second authors.}

\icmlcorrespondingauthor{Mahdieh Soleymani Baghshah}{soleymani@sharif.edu}

  % You may provide any keywords that you find helpful for describing your
  % paper; these are used to populate the "keywords" metadata in the PDF but
  % will not be shown in the document
  % \icmlkeywords{Machine Learning, ICML}

  \vskip 0.3in
]

% this must go after the closing bracket ] following \twocolumn[ ...

% This command actually creates the footnote in the first column listing the
% affiliations and the copyright notice. The command takes one argument, which
% is text to display at the start of the footnote. The \icmlEqualContribution
% command is standard text for equal contribution. Remove it (just {}) if you
% do not need this facility.

% Use ONE of the following lines. DO NOT remove the command.
% If you have no special notice, KEEP empty braces:
% \printAffiliationsAndNotice{}  % no special notice (required even if empty)
% Or, if applicable, use the standard equal contribution text:
\printAffiliationsAndNotice{\icmlEqualContribution}

\begin{abstract}
Large vision–language models (LVLMs) perform well on multimodal tasks, but their ability to reason and precisely align visual and textual information still has room for improvement. In this study, we show that external visual cues, such as symbols or grid lines, help LVLMs form more accurate connections between visual 
components, such as objects, and their corresponding textual descriptions,
improving their grounding and reasoning abilities. We introduce the concept of \textit{Grounding IDs}, which are latent identifiers 
that arise within the model as a result of external cues structuring both visual and textual modalities.
Our analysis reveals that partition-inducing external cues lead to Grounding IDs that make better alignment between corresponding visual and text representations, helping the model focus on relevant information. We find that Grounding IDs enhance attention between related components, improving cross-modal grounding and reducing hallucinations. 
Overall, our results show that Grounding IDs are a key mechanism that enables external cues to improve cross-modal alignment, reduce errors, and enhance the overall performance of LVLMs across a range of multimodal tasks.
\end{abstract}

\section{Introduction}

Large vision–language models (LVLMs), such as LLaVA \cite{llava}, GPT-4V \cite{gpt4}, and Qwen-VL \cite{qwen_vl}, have demonstrated strong performance on multimodal tasks like image captioning, visual question answering, and embodied tasks. %However, these models still face significant challenges in aligning visual and textual information accurately, leading to errors and hallucinations in generated text.
However, these models still face significant challenges in aligning visual and textual information accurately, leading to hallucinations in generated text and limited visual reasoning capabilities.
While recent studies, such as \citet{forgotten_polygons} and \textit{VISER} \cite{viser}, have shown that adding external structures like shape annotations or grid lines can improve performance, the mechanisms behind these improvements and their effects on model performance remain unclear. This presents a critical gap in understanding how LVLMs leverage these cues to reduce errors and enhance task performance.

% paragraph 3: Gap in interpretability, Our Hypothesis and Contribution: Grounding IDs, Overview of Methodology & Key Findings

To address this gap, we investigate the internal mechanisms of LVLMs under externally induced structure in the visual and textual modalities, revealing the emergence of \textit{Grounding IDs}. These are latent identifiers that the model generates to bind visual features, such as objects or spatial regions, to external cues like symbols or grid lines. Our central hypothesis is that when LVLMs are provided with such external structures, they create these Grounding IDs to improve the alignment between image and text. Through causal analyses, we demonstrate how these identifiers emerge within the model’s internal representations and how they propagate across embeddings. This finding clarifies how external structures improve the model's ability to link visual and textual information, enhancing both grounding and reasoning. Our work provides new insights into the internal mechanisms of LVLMs, contributing to a deeper understanding of how these models process and connect multimodal information.

%Beyond mechanistic interpretability, 
Furthermore, we extend prior work on simple scaffolds (e.g., horizontal lines)~\cite{viser} to more effective multimodal cues, which align input modalities through a set of unique marks incorporated in both visual and textual modalities. Our ablation studies show that both visual and textual cues contribute to binding, with their combination yielding the greatest improvements in cross-modal grounding. %We demonstrate the practical utility of Grounding IDs by showing that enhanced cross-modal attention reduces hallucinations in LVLMs, while also improving performance on visual reasoning tasks. Our results identify Grounding IDs as a key mechanism for partition-based binding, offering mechanistic insight into LVLM reasoning and delivering practical gains in both reducing hallucinations and enhancing visual reasoning capabilities.
We demonstrate the practical utility of \textit{Grounding IDs} by showing that enhanced cross-modal alignment reduces hallucinations in LVLMs and improves performance on visual reasoning tasks. Our results identify \textit{Grounding IDs} as a key mechanism for partition-based binding, providing mechanistic insight into LVLM when visual and textual modalities are structured by aligned external cues. Code is available at \url{https://github.com/sharif-ml-lab/GroundingID}.
% Our main contributions are as follows:

\begin{figure*}[t]
    \centering
    \includegraphics[width=0.90\linewidth]{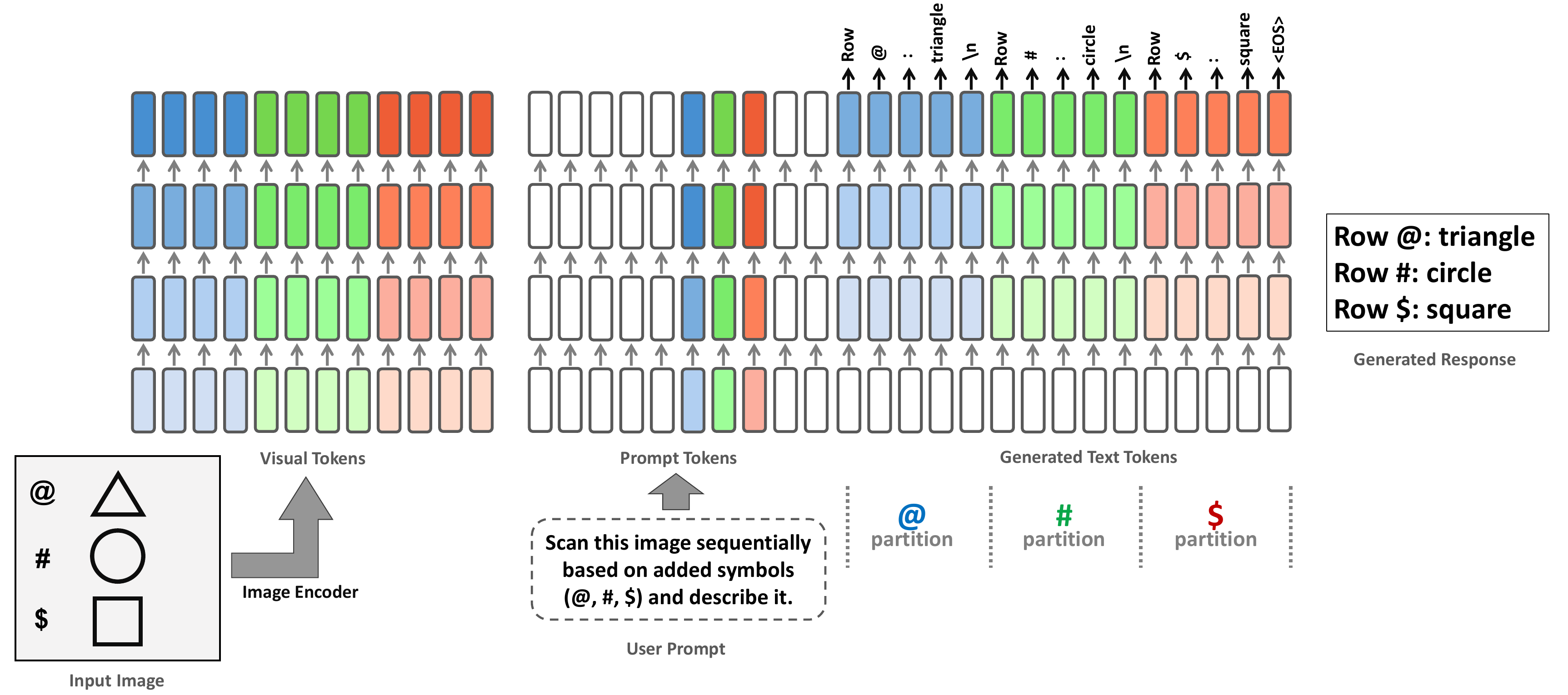}
    \vspace{-1mm}
    \caption{
    Conceptual overview of \textbf{Grounding IDs}. 
    An input image is augmented with simple visual cues (e.g., \{@, \#, \$\}) and paired with a prompt that explicitly includes these symbols. 
    Embeddings with the same Grounding IDs are displayed in matching colors across modalities, illustrating the reinforced binding between partitions and their corresponding textual descriptions.
    }
    \label{fig:graphical_abstract}
    \vspace{-1mm}
\end{figure*}

% % \vspace{-1.5mm}

\section{Background and Hypotheses}
% \section{The Grounding ID Hypothesis}

Earlier studies show that adding external artifacts, combined with chain-of-thought (CoT) style prompting, can substantially enhance the reasoning abilities of vision–language models by shifting them from one-pass perception toward more systematic, system-2-like processing. \citet{forgotten_polygons} demonstrated that LVLMs are often “shape-blind”: their vision encoders cluster frequent shapes but fail to distinguish less common ones, leading to errors in side counting and geometric reasoning. Explicit cues, such as annotated edges, encourage more deliberate strategies and yield large accuracy gains. 

\textit{VISER} \cite{viser} generalizes this idea beyond polygons by introducing versatile and input-agnostic visual structure, such as horizontal lines, paired with sequential scanning prompts that promote structured reasoning. This approach reduces feature binding errors, encourages serial scene parsing, and consistently improves performance across reasoning tasks like counting and visual search.
Building on this foundation, our work probes the internal circuits through which simple external artifacts shape visual reasoning in LVLMs.

Recent advances in mechanistic interpretability show that LLMs solve entity–attribute binding using \textbf{Binding IDs}, latent vectors that link entities with their attributes \cite{binding_id}. Causal mediation analyses support this finding and further show that these identifiers behave additively in representation space. Follow-up work extends this idea to vision–language models, showing that they form similar vectors connecting visual objects with textual references \cite{vision_binding_id}. Notably, this study is limited to very simple images where grounding is trivial, and issues such as information loss or cross-modal misalignment do not arise. %In contrast, these works examine binding between items and attributes in the standard setting. Inspired by these abstract identifiers, we investigate the underlying mechanism by which external visual structures, equipped with aligned textual cues, enhance LVLMs’ reasoning, particularly in more complex scenarios where grounding is non-trivial and cross-modal misalignment is more pronounced.
These works examine binding between items and attributes in the standard setting. However, we intend to investigate the underlying mechanism by which external visual structures, equipped with aligned textual cues, enhance LVLMs’ reasoning, particularly in more complex scenarios where grounding is non-trivial and cross-modal misalignment is more pronounced.

In this work, we aim to answer a key open question: \textit{why do external cues improve reasoning in LVLMs?} To this end, we introduce a general setting where both images and prompts are augmented with simple shared cues (e.g., symbol characters) that partition the input into distinct regions. With these aligned multimodal cues, the models appear to generate abstract identifiers that bind objects to their respective partitions, supporting more systematic visual scanning.
Fig.~\ref{fig:graphical_abstract} illustrates how these identifiers emerge both in representations of partitioned areas and in the generated textual descriptions of those partitions. We refer to these identifiers as Grounding IDs, since this partition-based binding enhances multimodal grounding.
In particular, this study seeks to validate the following hypotheses:
\begin{itemize}
\item
\textbf{Existence}: Augmenting the inputs with multimodal external cues induces Grounding IDs that propagate through embeddings and attention, establishing within-partition binding across modalities.
\item
\textbf{Modality Gap}: Grounding IDs reduce the alignment gap between image and text representations of corresponding tokens.
\end{itemize}

\begin{figure*}[htb]
    \centering

    \begin{subfigure}{0.19\textwidth}
        \includegraphics[width=\linewidth]{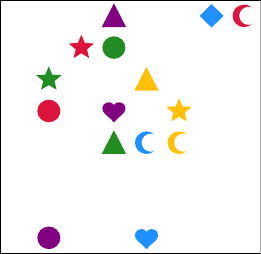}
    \end{subfigure}
    \hspace{0.01\textwidth}
    \begin{subfigure}{0.20\textwidth}
        \includegraphics[width=\linewidth]{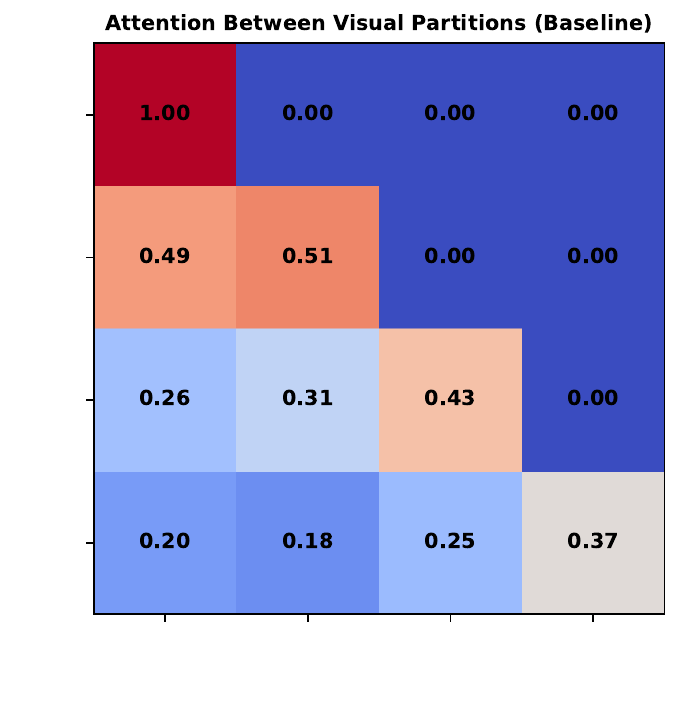}
        \vspace{-5.5mm}
    \end{subfigure}
    \hspace{0.01\textwidth}
    \begin{subfigure}{0.20\textwidth}
        \includegraphics[width=\linewidth]{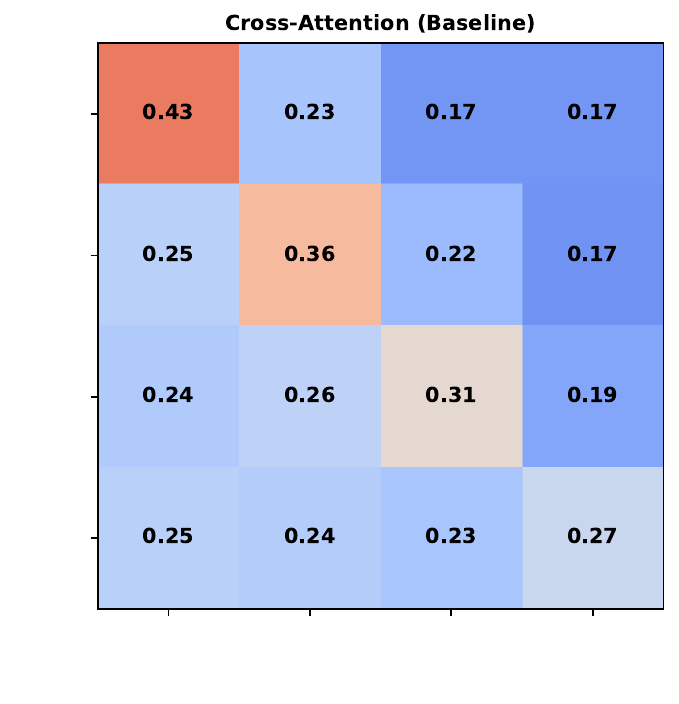}
        \vspace{-5.5mm}
    \end{subfigure}

    \vspace{0.1cm}

    \begin{subfigure}{0.19\textwidth}
        \includegraphics[width=\linewidth]{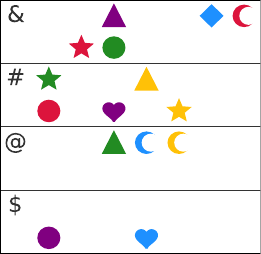}
        \caption{}
    \end{subfigure}
    \hspace{0.02\textwidth}
    \begin{subfigure}{0.19\textwidth}
        \includegraphics[width=\linewidth]{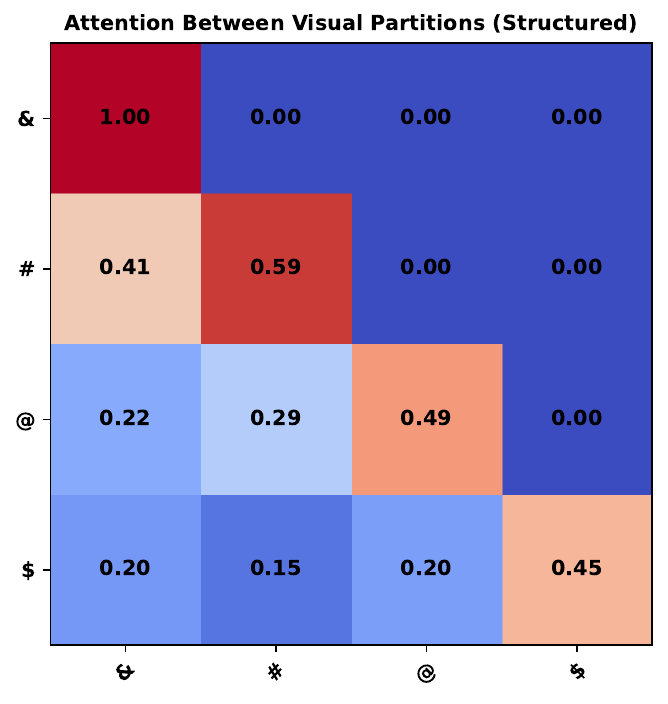}
        \vspace{-6mm}
        \caption{}
    \end{subfigure}
    \hspace{0.02\textwidth}
    \begin{subfigure}{0.19\textwidth}
        \includegraphics[width=\linewidth]{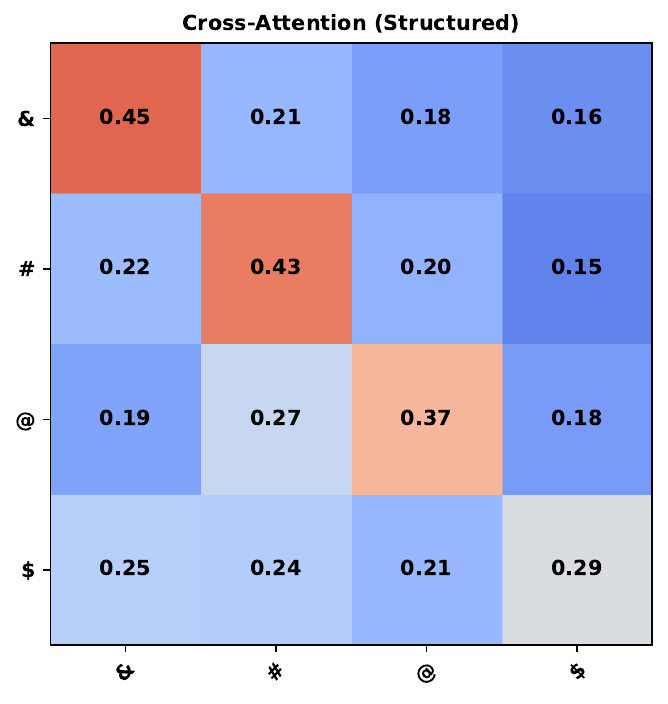}
        \vspace{-6mm}
        \caption{}
    \end{subfigure}
    \vspace{-1mm}
    \caption{
    Illustration of attention patterns under baseline and structured inputs in the scene description task. 
    (a) One dataset sample (top: baseline, bottom: structured). 
    (b) Within-modality visual attention matrices. 
    (c) Cross-modality attention matrices. 
    Values are averaged over 500 samples and layers 22--27 of Qwen2.5-VL.
    }
    \label{fig:attention_mat}
    \vspace{-1mm}
\end{figure*}

To evaluate these hypotheses, the remainder of the paper is organized as follows. In Section~\ref{sec:existence}, we probe model representations and internal dynamics to provide empirical evidence for Grounding IDs, showing how they emerge and influence attention patterns and cross-modal alignment. Section~\ref{sec:causality} presents causal intervention experiments that investigate how Grounding IDs contribute to object–cue binding. Finally, Section~\ref{sec:hallucination} explores the practical implications, demonstrating that enhancing cross-modal binding through Grounding IDs reduces hallucinations in LVLMs.

% These analyses provide converging evidence that external cues induce Grounding IDs, which serve as a core mechanism for partition-based binding in LVLMs.

%\section{Empirical Evidence for Existence of Grounding IDs} 
\section{Evidence of Improved Alignment}
\label{sec:existence}

In this section, we provide empirical evidence for the existence of Grounding IDs by analyzing the internal representations and attention patterns of LVLMs. Our goal is to investigate whether inducing structure with external cues, such as symbols and lines, improves the alignment between visual and textual components. We focus on inference-time reasoning using a 7B Qwen2.5-VL model without fine-tuning. The tasks involve scene description and visual question answering, where the model generates detailed descriptions of the scene based on the provided input. Results for additional models are provided in Appendix~\ref{sec:extended_experiments}.

We compare two setups: a baseline and a structured input method. The baseline uses an unmodified image with a standard scene-description prompt. In contrast, the structured input method augments the baseline by adding four symbols (\&, \#, \$, @) to the image and prompt, and dividing the image into four horizontal partitions with three lines, as shown in Fig.~\ref{fig:attention_mat}(a). This structured input is designed to provide additional cues that guide the model in better aligning visual and textual information. To assess the impact of different cues, we also conduct ablation studies with alternative cue designs, which are detailed in Appendix~\ref{sec:cue:ablation}.

The analysis is conducted on a synthetic dataset of images with varying object configurations. Each image contains 15 unique objects drawn from 35 shape–color combinations (7 shapes × 5 colors), with each object occupying a single patch (28×28 pixels) and not extending into adjacent patches. After the model generates its output, we match textual tokens to visual objects using regular-expression (regex) pattern matching, as each object type appears only once in the image. This process assigns a partition label to each token in both modalities, %allowing us to compute attention and embedding statistics within specific partitions.
allowing us to compute partition-wise attention and embedding statistics. 
We then analyze attention patterns and embedding similarities both within individual modalities and across the visual-textual interface. %These analyses provide empirical evidence that the alignment between visual and textual components improves with the introduction of structured cues, suggesting that the model may have a mechanism that leads to the emergence of Grounding IDs, which help the model more effectively linking objects and their corresponding descriptions.
These analyses provide empirical evidence that structuring the image and generated text improves alignment between visual and textual components, suggesting the presence of a mechanism that more effectively links objects with their corresponding descriptions.

\begin{figure}[htb]
    \centering
    \begin{subfigure}{\linewidth}
        \centering
        \includegraphics[width=0.80\linewidth]{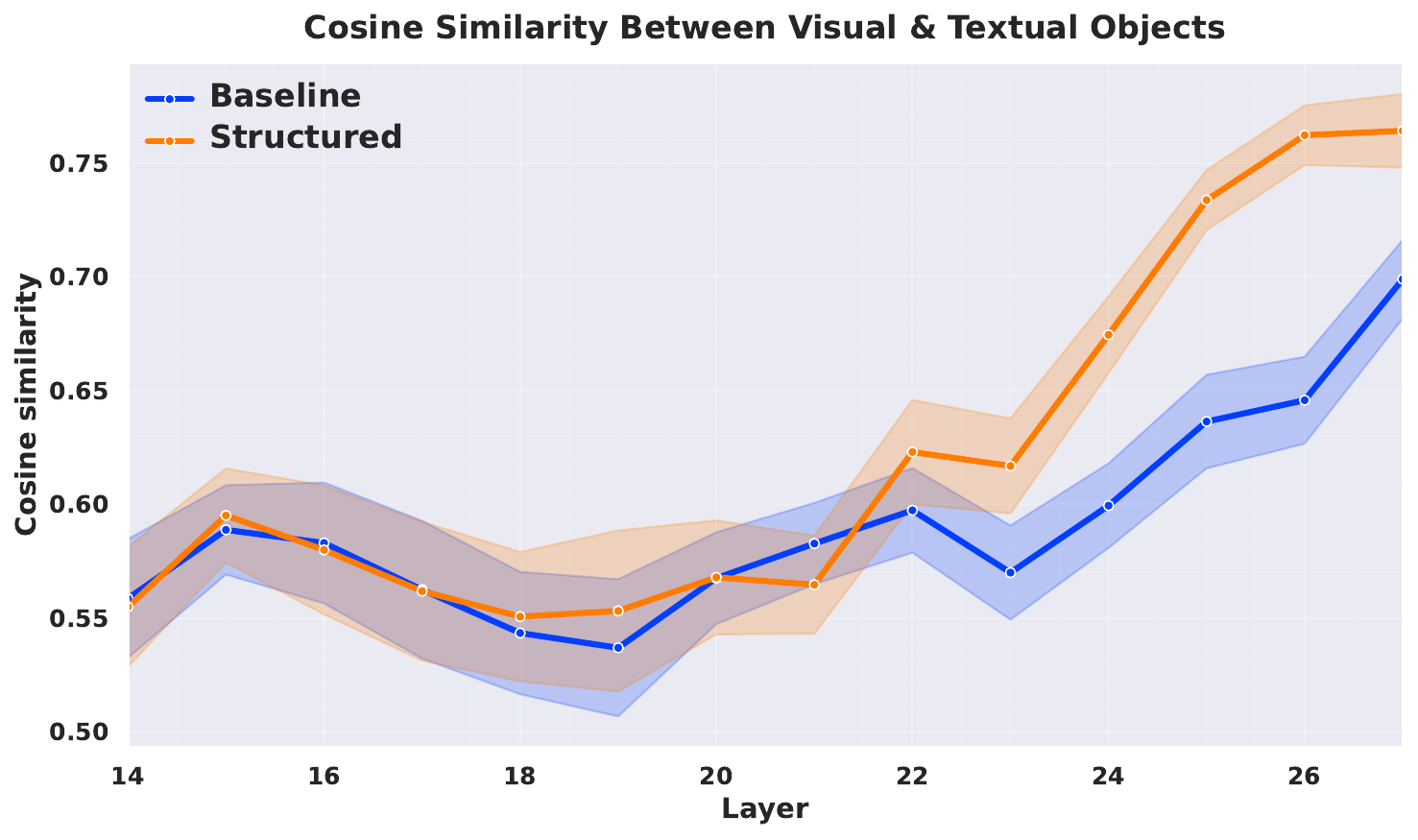}
        \vspace{-1mm}
        \caption{}
        \label{fig:modality_gap_layers}
    \end{subfigure}
    \begin{subfigure}{0.62\linewidth}
        \centering
        \includegraphics[width=\linewidth]{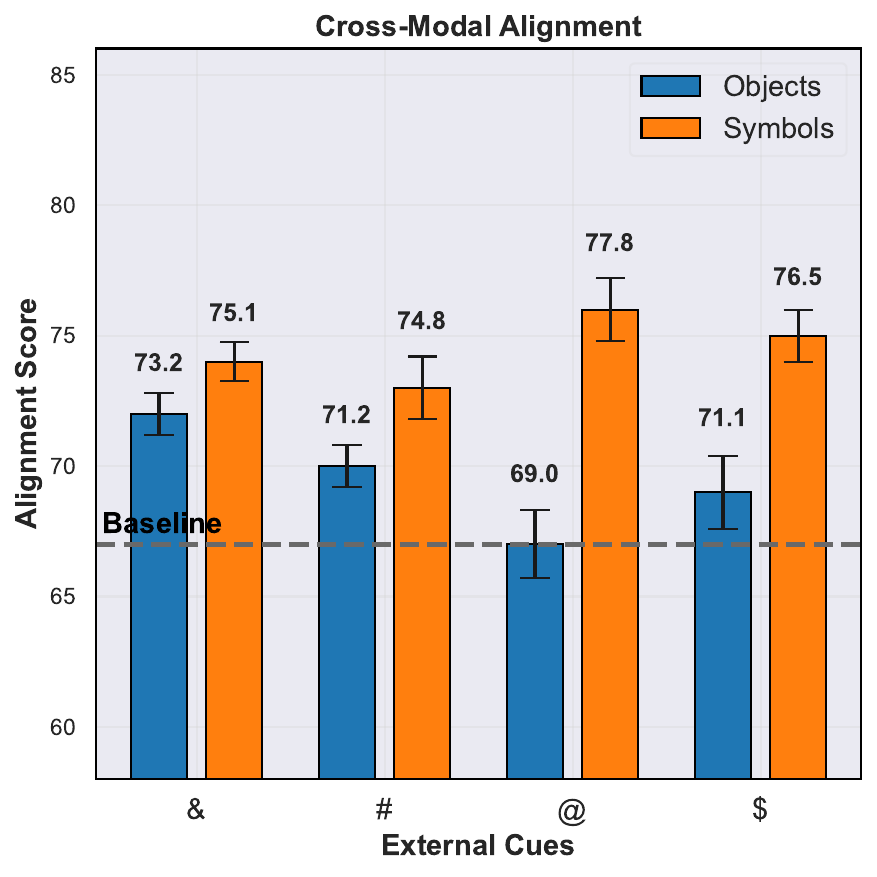}
        \vspace{-1mm}
        \caption{}
        \label{fig:modality_gap_average}
    \end{subfigure}
    \vspace{-1mm}
    \caption{
    Analysis of the modality gap. 
    (a) Cross-modal alignment across layers,
    % averaged over 100 samples,
    showing that improvements emerge in layers 22--27. 
    (b) Average alignment in layers 22--27, reported separately for four partitions. 
    Object embeddings under structured inputs achieve higher alignment than the baseline (dashed line), and symbol embeddings achieve even stronger alignment than objects.
    }
    \label{fig:alignment_mat}
    \vspace{-1mm}
\end{figure}

\paragraph{Attention Analysis.}

For each token, we take the maximum attention score over all heads and then average across tokens within each partition (image rows), yielding a $4 \times 4$ matrix. Aggregation is performed on true positive objects, where the model generates descriptions, and the corresponding objects are present in the image. Descriptions (including shape and color) are linked to the correct objects, ensuring accurate attention. To avoid ambiguity, the shapes in the image are unique, ensuring proper association between descriptions and objects. The aggregated attention score is averaged across true positives, providing a measure of cross-modal alignment between the model’s generated text and the objects in the image. Fig.~\ref{fig:attention_mat} shows both within-modality and cross-modality attention matrices for baseline and structured inputs. Despite the simplicity of the external cues, structured inputs exhibit stronger diagonal dominance, with attention concentrated within partitions. See also Appendix~\ref{sec:att_1} and~\ref{sec:att_2} for complementary attention analysis.

\paragraph{Modality Gap.}
While attention captures binding structure, embedding similarity provides a complementary view of alignment between modalities. We measure cosine similarity between visual and textual embeddings corresponding to correctly generated object tokens. Similarities are computed layer-wise and averaged across samples. Both the baseline and the structured case show alignment strengthening in later layers (after layer 20). Structured modalities consistently achieve higher similarity, particularly in the last four layers, as shown in Fig.~\ref{fig:alignment_mat}. This confirms that external cues reduce the modality gap in LVLMs by enhancing cross-modal alignment. In particular, cosine similarity between activation patches corresponding to external symbols across modalities is higher than that of dataset objects. 
%\citet{modality_gap} report that post-hoc removal of gap-contributing dimensions, while effective in reducing the modality gap in VLMs, distorts the embedding geometry and degrades performance. In contrast, our results show that external multimodal cues reduce the modality gap through the model’s internal dynamics without such disruptive interventions.

\begin{figure}[t]
    \centering
    \begin{subfigure}{0.96\linewidth}
        \centering
        \includegraphics[width=\linewidth]{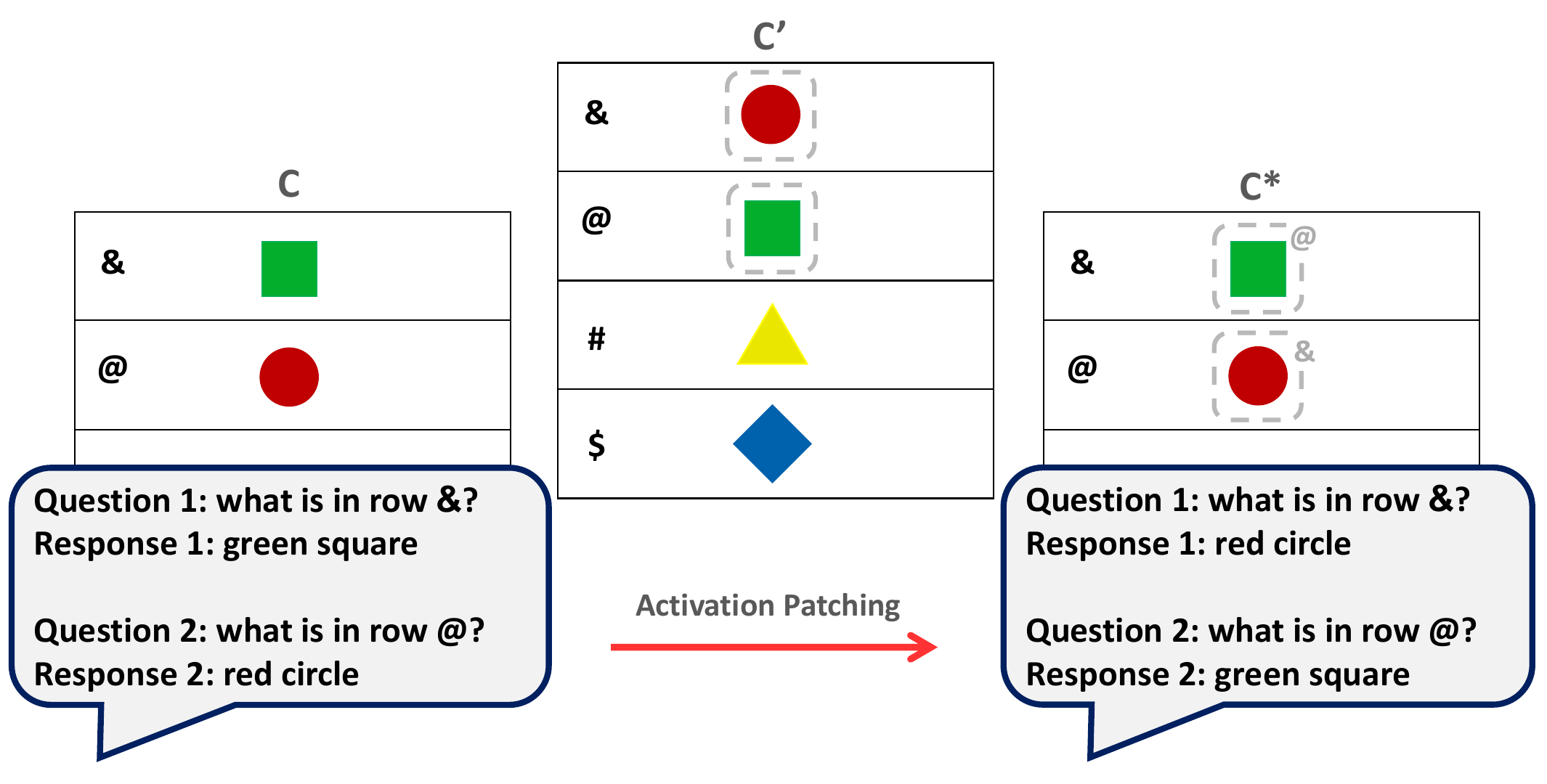}
        \vspace{-2mm}
        \caption{}
        \label{fig:swap_example}
    \end{subfigure}
    \begin{subfigure}{0.74\linewidth}
        \centering
        \includegraphics[width=\linewidth]{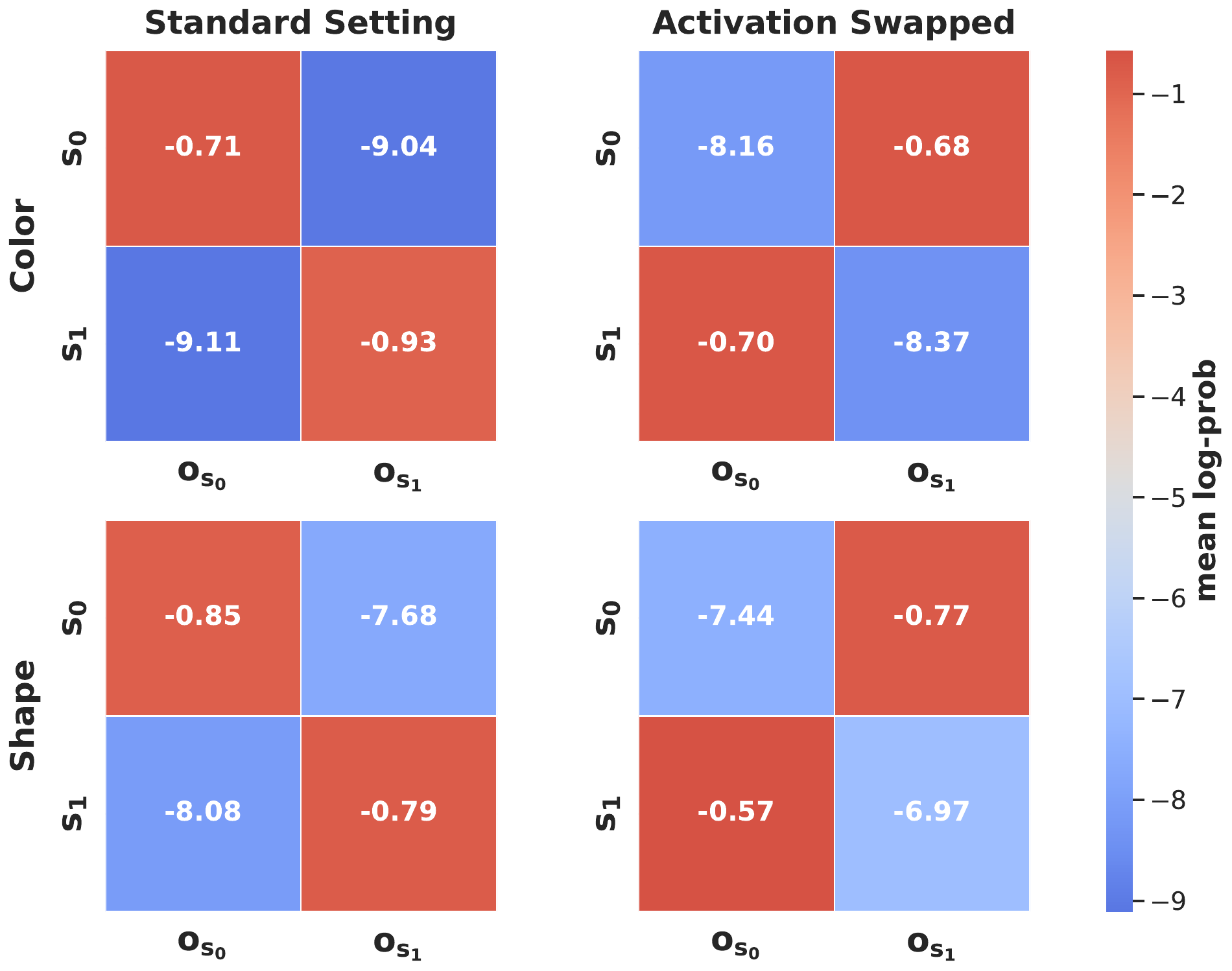}
        \vspace{-2mm}
        \caption{}
        \label{fig:swap_result}
    \end{subfigure}
    \vspace{-1mm}
    \caption{Activation swap experiment.
    (a) Procedure in a case where source ($c'$) and target ($c$) contain the same objects. 
    Activations from the \& and @ partitions of $c'$ are patched into $c$, producing the patched context $c^*$. Predictions in $c^*$ follow the transferred bindings (gray) rather than host symbols.
    (b) Average log probabilities of $c$ and $c^*$ over valid row–symbol–object combinations. Rows and columns indicate the two selected query symbols and their corresponding objects.
    }
    \label{fig:swap_intervention}
    \vspace{-2mm}
\end{figure}

\section{Causal Evidence of Grounding IDs} \label{sec:causality}

The alignment and attention analyses in Section~\ref{sec:existence} provide correlational evidence that external cues induce partition-based alignment. Our hypothesis is that \textit{Grounding IDs} act as abstract vectors that are induced to related tokens across modalities, enhancing multimodal binding and thereby increasing alignment and grounding. We now seek causal validation using a medium-sized LVLM, Qwen2.5-VL 7B (see Appendix~\ref{sec:extended_experiments} for additional models). For causal mediation analysis, we use a simplified synthetic dataset in which each image contains four rows with one object per row. Rows are labeled with non-ordinal symbols \{\&,\$,\#,@\} to prevent sequential order cues.
The task is discriminative visual question answering: the prompt asks, for example, ``What is in row @?'', and the model must respond with the correct object description, such as ``red circle.''

To formalize notation, we define $\mathbf{o}^{s_j}_{s_i}$
as an object $\mathbf{o}$ that is \emph{located} in the partition associated with symbol $s_i$ (subscript) and \emph{bound} by the model to symbol $s_j$ (superscript).
For example, $\mathbf{o}^{\$}_{\&}$ denotes an object positioned in the \& partition but bound to \$.
For convenience, we use two shorthand cases:

\textbullet \hspace{0.3em} $\mathbf{o}^{\sim s}_{s}$: an object located in the partition of $s$ but bound to a different symbol than $s$.

\textbullet \hspace{0.3em} $\mathbf{o}^{s}_{\sim s}$: an object bound to $s$ but located in a different partition than $s$.

\subsection{Activation Swapping Experiment}

We follow a causal mediation framework similar to the introduced one in prior studies on language models \cite{vig2020investigating, binding_id}. We adopt standard terminology from mechanistic interpretability to describe the causal experiments in this section~\cite{MI_Survey}.
We denote by $c$ an input \emph{context} consisting of an image $x$ and a text prompt $p$. Given a model $M$ and layer $\ell$, let $h^{(\ell)}(c)$ be the hidden activations (residual stream) at that layer, including both visual patch tokens and textual tokens.
In our interventions, we use three contexts: a \emph{target} context $c$, a \emph{source} context $c'$, and a \emph{patched} context $c^*$ obtained by replacing a subset of activations in $h^{(\ell)}(c)$ with those from $h^{(\ell)}(c')$ and then running the remaining layers and tokens of $M$ on the patched context $c^*$.

For our analysis, two contexts \( c \) and \( c' \) are randomly sampled from the controlled dataset, and their activations are extracted across all layers (see Appendix~\ref{app:act-swap} for details on the dataset). We then select two random symbols (e.g., \& and @) and swap the patch activations corresponding to their objects between the two contexts. Specifically, the activations from all layers of the object in row \& of \( c' \) are replaced by the activations of the corresponding patches in row @ of \( c \), while the object activations in row @ of \( c' \) are patched into row \&. We pass the patched input \( c^{*} \) to the model and record its predictions for the selected symbols. Fig.~\ref{fig:swap_example} shows a simple case where the swapped objects are originally assigned different symbols between \( c \) and \( c' \). This swap strategy is designed to avoid ambiguity that could arise from objects being bound to the same symbol in the patched context (i.e., coexisting \(\mathbf{o}^{s}_{s}\) and \(\mathbf{o}^{s}_{\sim s}\)).

%of two identical objects in the image  

%By analyzing the predictions of the patched context $c^{*}$, we observe that the model consistently follows the binding symbols of the swapped objects rather than the symbols present locally in the host context. In other words, for a given partition $s$, the model outputs $\mathbf{o}^{s}_{\sim s}$ instead of $\mathbf{o}^{\sim s}_{s}$ (adjacent object).
By analyzing the predictions of the patched context $c^{*}$, we observe that the model consistently follows the symbols bound to the objects in the source rather than the symbols present locally in the host context.
For example, if the object in row~@ of $c'$ is a green square and it is inserted into row~\& of $c$, the model reports a green square for row~@, even though that object is physically located beside \&. This indicates that the symbol associated with a partition is encoded in the embeddings of objects belonging to that partition and is transferred to the patched context through the patched objects. Moreover, the model disregards local cues in the patched context $c^*$ and follows the transferred symbol-object binding of $c'$ for its final prediction (Fig.~\ref{fig:swap_example}).

To quantify this effect across the dataset, we consider all valid swaps, where object patches are swapped between two contexts, $c$ and $c'$, ensuring that the objects differ in both shape and color between the contexts. We evaluate the model on the patched context, $c^*$, using a fixed set of query symbols $s \in \mathcal{S}$. If the model’s answer for symbol $s$ in $c^*$ matches $\mathbf{o}^{s}_{\sim s}$ (i.e., it follows the transferred binding rather than the locally present object $\mathbf{o}^{\sim s}_{s}$), we count the trial as correct. Swap accuracy is the fraction of such queries for which the model outputs $\mathbf{o}^{s}_{\sim s}$. For further details on how valid swaps are generated and the full procedure, please refer to Section~\ref{app:act-swap} in the appendix.

If we follow the standard setting without intervention and treat %$\mathbf{o}^{(\cdot)}_{s}$ as the class label
 object located in the partition corresponding to symbol $s$
as the desired output, standard accuracy drops sharply from $\mathbf{1.00}$ to $\mathbf{0.02}$ after swapping. In contrast, if we treat %$\mathbf{o}^{s}_{\sim s}$ as the correct label (i.e., prediction based on transferred bindings),
the object bound to symbol s in the target context being transferred to the patched context as the desired output,  
swap accuracy remains high at $\mathbf{0.98}$, confirming that predictions follow Grounding IDs. Thus, object-symbol bindings are causally mediated by these IDs. Fig.~\ref{fig:swap_result} shows the average log-probability of predicted objects before and after activation swapping.
Additional intervention results on more complex synthetic layouts and natural images are provided in Appendix~\ref{app:extended_intervention}, showing that the effect persists beyond the single-object-per-row setting.

\begin{figure}[t]
    \centering
    \begin{subfigure}{0.77\linewidth}
        \centering
        \includegraphics[width=\linewidth]{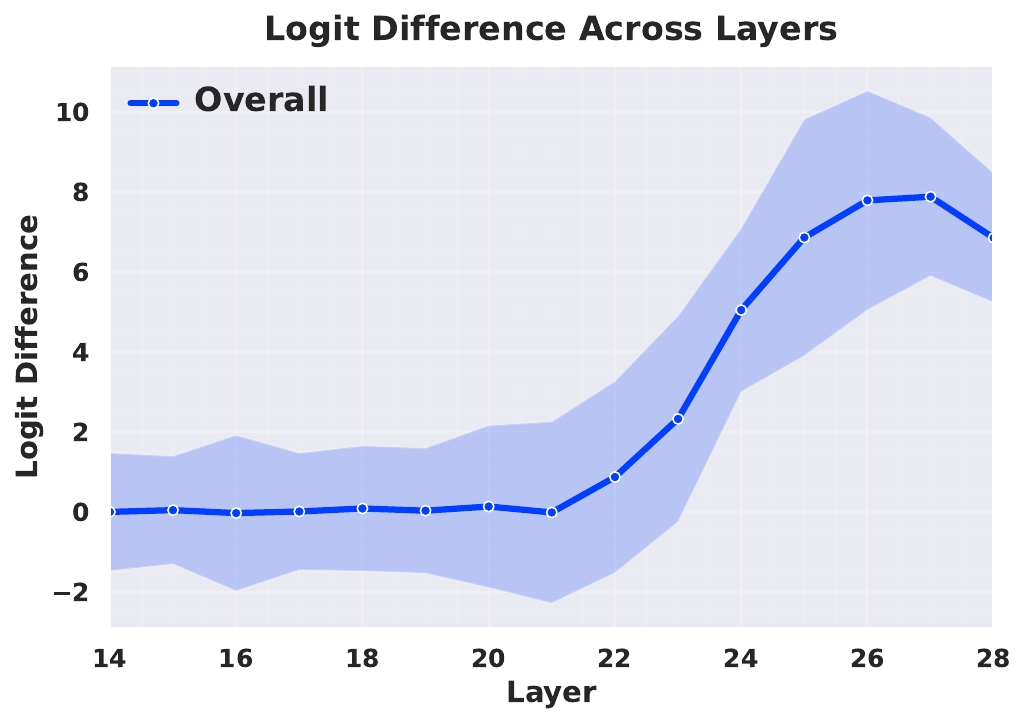}
        \caption{}
        \label{fig:intervention:logit_lens}
    \end{subfigure}
    \hfill
    \begin{subfigure}{0.80\linewidth} 
        \centering
        \includegraphics[width=\linewidth]{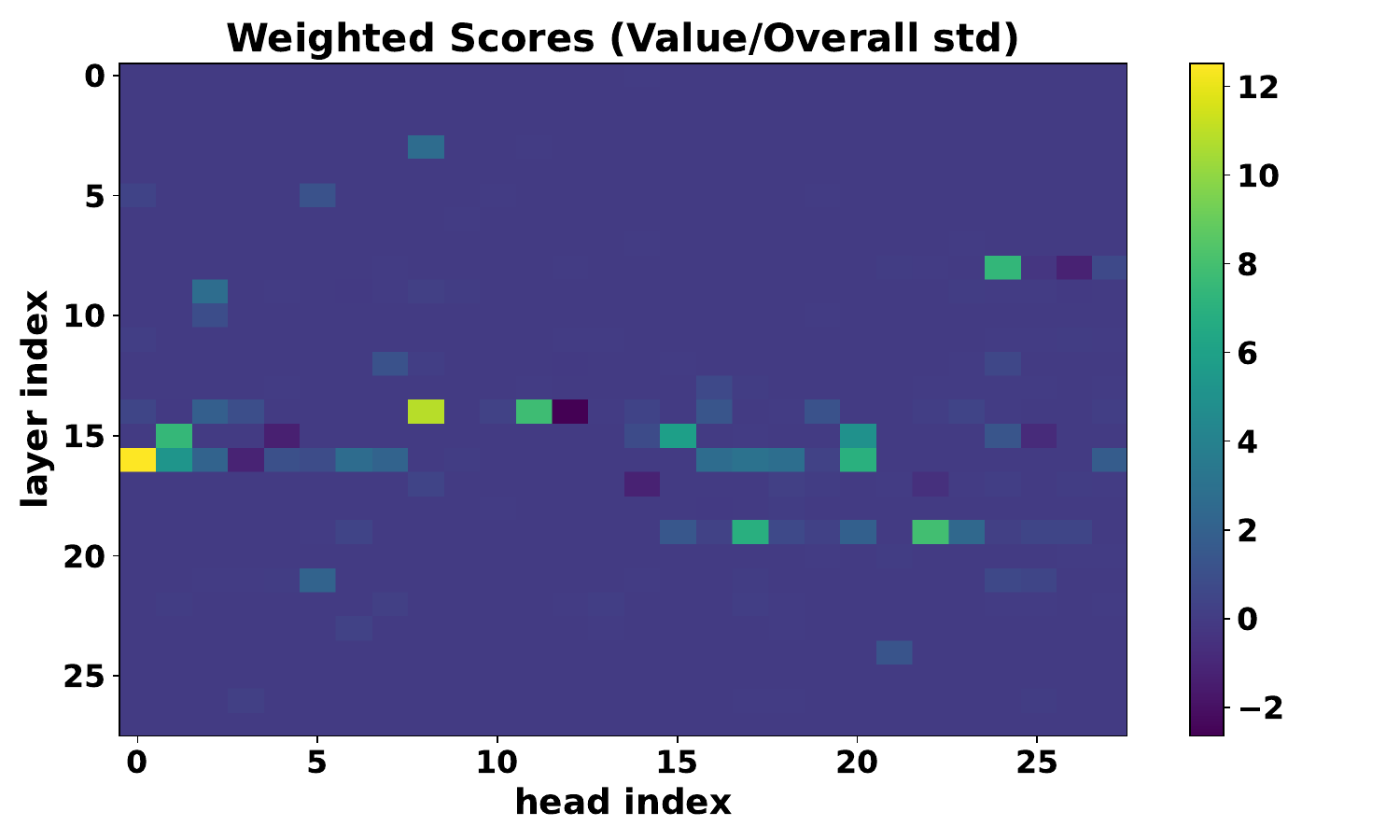}
        \caption{}
        \label{fig:intervention:attention_heatmap}
    \end{subfigure}
    \vspace{-1mm}  % Reduce space between subfigures and the caption
    \caption{
    Causal mediation analysis across layers.  
    (a) Average logit differences between $\mathbf{o}^{s}_{\sim s}$ and $\mathbf{o}^{\sim s}_{s}$ across layers, showing where the model begins to favor the bound object $\mathbf{o}^{s}_{\sim s}$.
    (b) Signal-to-noise scores for attention differences between $\mathbf{o}^{s}_{\sim s}$ and $\mathbf{o}^{\sim s}_{s}$ patches across heads and layers.
    }
    \label{fig:intervention_layerwise}
    \vspace{-1mm}  % Reduced space after the caption
\end{figure}

\subsection{Layerwise Analysis of Grounding ID Emergence}

\textbf{Layerwise logit difference.}
Here, we investigate in which layers the model begins to predict the intervened $\mathbf{o}^{s}_{\sim s}$ (bound object to $s$) rather than $\mathbf{o}^{\sim s}_{s}$ (adjacent object to $s$). The goal is to identify the layer at which the model’s prediction shifts toward the object carrying the transferred binding in the patched context. In this experiment, we use a monochrome dataset for simplicity. We apply the logit lens technique \cite{nostalgebraist20_logit_lens} 
to the single-word response token at each layer~$\ell$.
Let $L^{(\ell)}(x \mid c^*)$ be the unnormalized logit for token $x$ decoded from $h^{(\ell)}(c^*)$.
We define the layerwise logit difference
\begin{equation}
\Delta L^{(\ell)}
\;=\;
L^{(\ell)}\!\bigl(\mathbf{o}^{s}_{\sim s} \mid c^*\bigr)
\;-\;
L^{(\ell)}\!\bigl(\mathbf{o}^{\sim s}_{s} \mid c^*\bigr),
\end{equation}
to compute the prediction tendency between $\mathbf{o}^{s}_{\sim s}$ and $\mathbf{o}^{\sim s}_{s}$ in each patched context, and then average across all valid symbol–object pairs.
Positive values of $\Delta L^{(\ell)}$ indicate that, at layer~$\ell$, the representation favors the bound object over the local object.

As shown in Fig.~\ref{fig:intervention:logit_lens}, this difference becomes positive in the later layers (20–27), indicating that the model increasingly favors the bound object after the intervention. This result complements the representational trend in Fig.~3a, where alignment for structured inputs increases in the same higher layers, showing that the late-layer alignment shifts are accompanied by causal evidence of Grounding ID usage.

{\textbf{Responsible attention heads.}} We also investigate which attention heads are most responsible for propagating Grounding IDs as opposed to relying on local visual proximity. For each head and layer, we compute the difference in attention from the response token to visual tokens corresponding to {$\mathbf{o}^{s}_{\sim s}$ vs. $\mathbf{o}^{\sim s}_{s}$ (the bound object vs. the adjacent one)} across samples. To quantify consistency, we divide the mean difference by its standard deviation, yielding a signal-to-noise ratio that reflects how reliably a head prefers bound objects.
To emphasize attention to meaningful regions, this ratio is multiplied by the head’s average attention weight.
Let $\alpha^{(\ell,h)}(\mathbf{r} \!\to\! \mathbf{o})$ denote %the attention weight from the response token $\mathbf{r}$ to the patch of object $\mathbf{o}$
the attention weight of the response token $\mathbf{r}$ on the patch corresponding to object $\mathbf{o}$
at layer~$\ell$, head~$h$.
For each head, we compute
% \begin{equation}
% S^{(\ell,h)}
% \;=\;
% \frac{
%     \mathbb{E}\!\left[
%         \alpha^{(\ell,h)}(\mathbf{r}\!\to\!\mathbf{o}^{s}_{\sim s})
%         -
%         \alpha^{(\ell,h)}(\mathbf{r}\!\to\!\mathbf{o}^{\sim s}_{s})
%     \right]
% }{
%     \operatorname{Std}\!\left[
%         \alpha^{(\ell,h)}(\mathbf{r}\!\to\!\mathbf{o}^{s}_{\sim s})
%         -
%         \alpha^{(\ell,h)}(\mathbf{r}\!\to\!\mathbf{o}^{\sim s}_{s})
%     \right]
% }
% \cdot
% \mathbb{E}\!\left[\alpha^{(\ell,h)}(\mathbf{r}\!\to\!\text{image})\right],
% \end{equation}

\begin{equation}
\begin{aligned}
S^{(\ell,h)}
&=
\frac{
    \mathbb{E}\!\left[
        \alpha^{(\ell,h)}(\mathbf{r}\!\to\!\mathbf{o}^{s}_{\sim s})
        -
        \alpha^{(\ell,h)}(\mathbf{r}\!\to\!\mathbf{o}^{\sim s}_{s})
    \right]
}{
    \operatorname{Std}\!\left[
        \alpha^{(\ell,h)}(\mathbf{r}\!\to\!\mathbf{o}^{s}_{\sim s})
        -
        \alpha^{(\ell,h)}(\mathbf{r}\!\to\!\mathbf{o}^{\sim s}_{s})
    \right]
}
\\
&\quad\cdot
\mathbb{E}\!\left[\alpha^{(\ell,h)}(\mathbf{r}\!\to\!\text{image})\right].
\end{aligned}
\end{equation}

where expectations are taken over samples and valid symbol–object pairs. Heads with high $S^{(\ell,h)}$ consistently prefer bound objects over merely co-located ones and are interpreted as key carriers of Grounding IDs.
The resulting scores are visualized in Fig.~\ref{fig:intervention:attention_heatmap}, showing that Grounding IDs cause significant attention shifts in certain heads, particularly in middle layers (layer 16), which attend more to bound objects.

\subsection{The Characteristics of Grounding IDs}

% Activation swap with disjoint symbols

\begin{figure}[t]
    \centering

    \begin{subfigure}{0.83\linewidth}
        \centering
        \includegraphics[width=\linewidth]{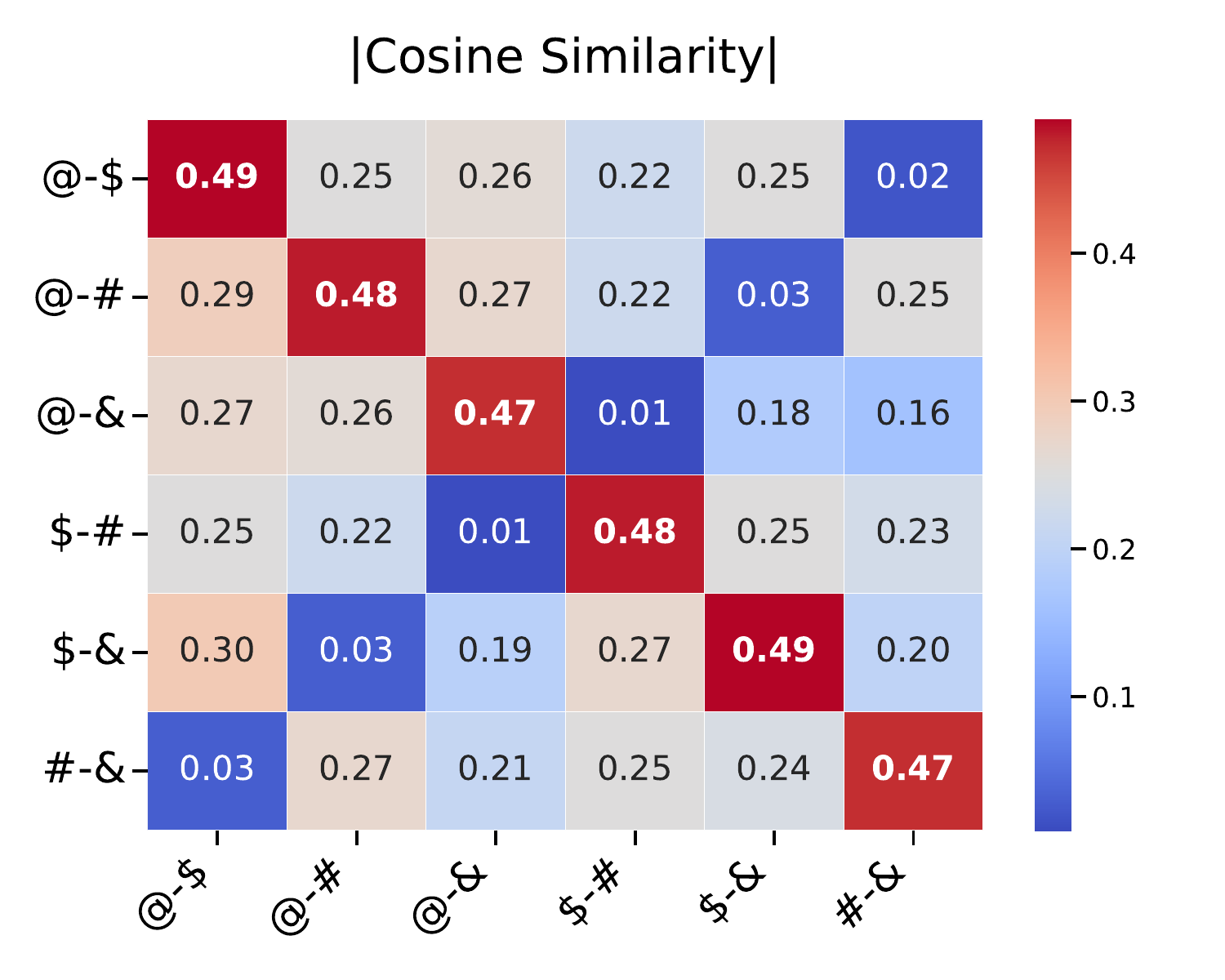}
        \vspace{-3mm}
        \caption{}
        \label{fig:grounding_alignment}
        % \vspace{-1mm}
    \end{subfigure}
    \begin{subfigure}{0.85\linewidth}
        \centering
        \begin{subfigure}{0.42\linewidth}
            \centering
            \includegraphics[width=\linewidth]
            {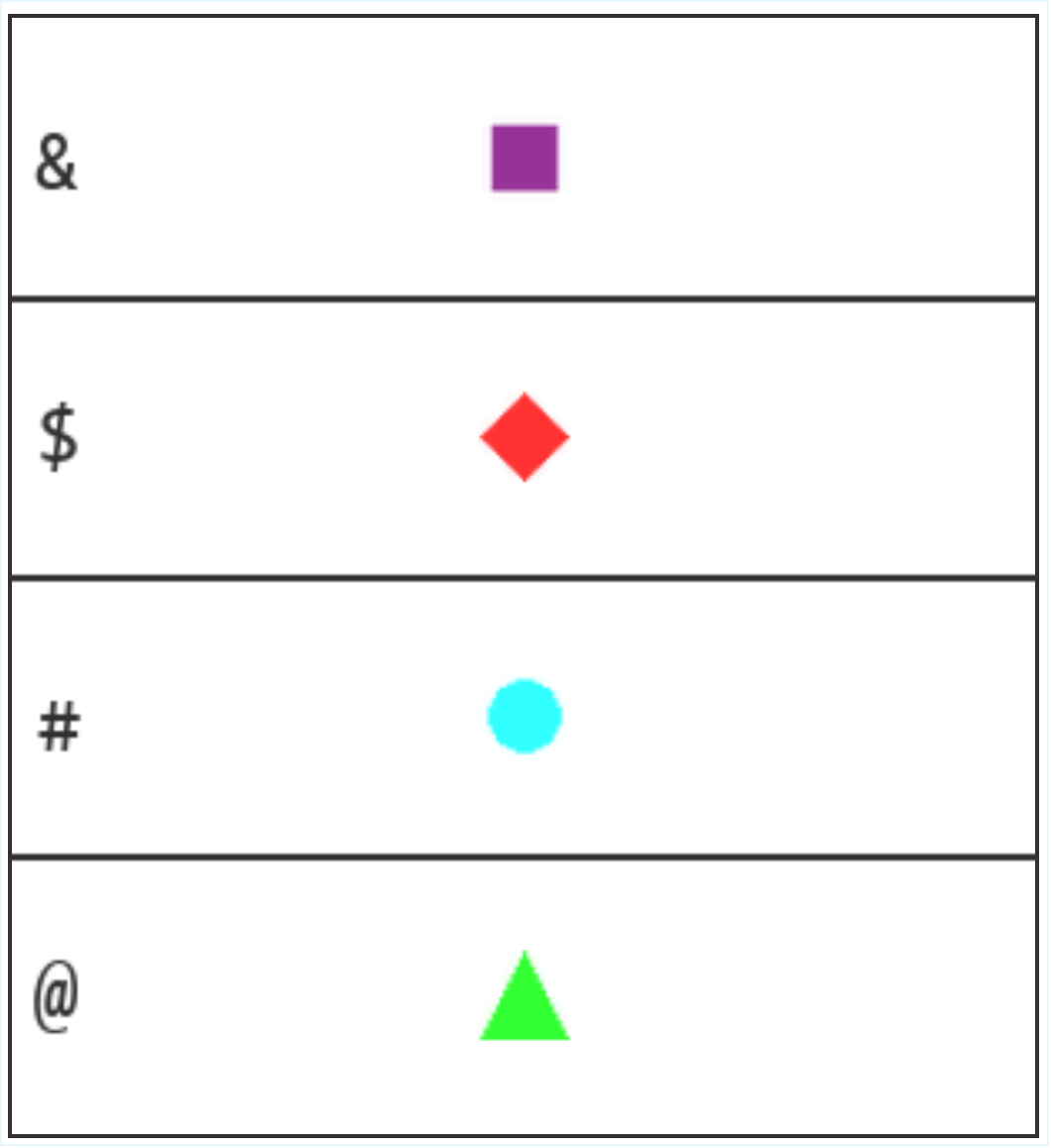}
        \end{subfigure}
        \hfill
        \begin{subfigure}{0.42\linewidth}
            \centering
            \includegraphics[width=\linewidth]{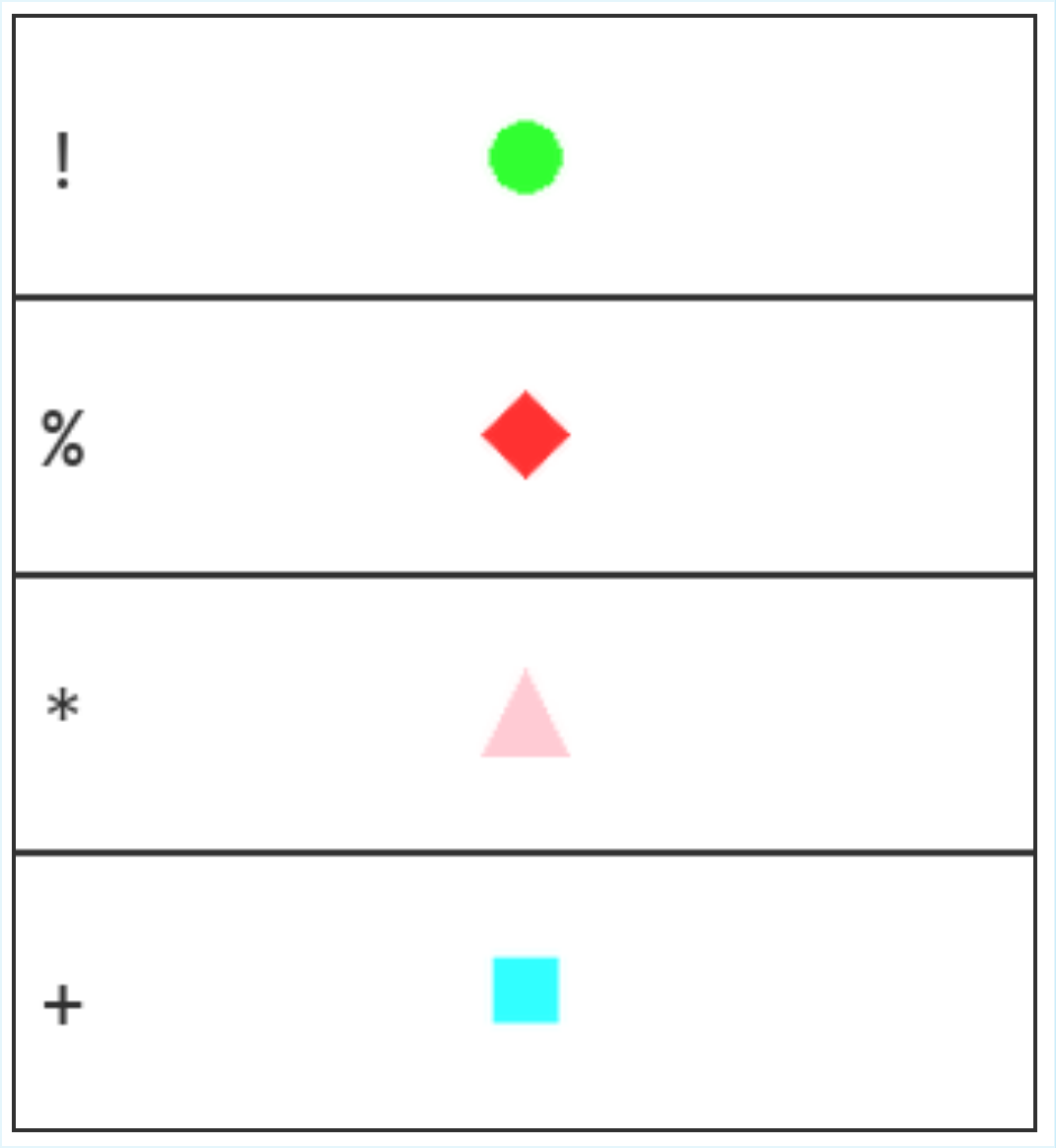}
        \end{subfigure}
        \vspace{-2mm}
        \caption{}
        \label{fig:disjoint_swap_stimuli}
        \vspace{-1mm}
    \end{subfigure}

    \caption{
        Characteristics of Grounding IDs.
        (a) Cosine similarity between averaged differential vectors of symbol patches (vertical) and their corresponding Grounding IDs (horizontal).
        (b) Disjoint-symbol control for the activation-swap test. The source context contains the symbols \{\&,\$,\#,@\}, while the target context contains a nonoverlapping set \{!,\%,$\times$, +\}.
    }
    \label{fig:grounding_id_characteristics}
    \vspace{-1mm}
\end{figure}

\citet{binding_id} shows that, in LLMs, Binding IDs linking entities and their attributes are largely context independent. In contrast, our causal mediation experiments demonstrate that Grounding IDs are directly predictable from their corresponding symbols, suggesting a mechanism closer to lexical binding~\cite{mixing_mechanisms}. To probe the characteristics of Grounding IDs, we analyze the relational similarity between symbols and their induced Grounding IDs. Concretely, for each pair of symbols (e.g., \&, \#), we compute the difference between their symbol patch activations. For the same pairs, we also compute the difference between the corresponding object patch activations. Averaging these differential vectors across the dataset cancels out confounding factors such as shape and position, yielding two structured spaces: one defined by symbol differences and one by Grounding ID differences. Fig.~\ref{fig:grounding_alignment} reports cosine similarities between these two spaces, revealing a strong correspondence between the symbol space and the Grounding ID space, which further supports a lexical-style binding mechanism.

To further assess this conjecture, we repeat the activation swap experiment but assign the target image a disjoint set of symbols (e.g., ${+, \times, \%, !}$) that do not overlap with those in the source (see Fig.~\ref{fig:disjoint_swap_stimuli} and Appendix~\ref{sec:disjoint_swap} for details). After activation patching, we query the model in $c^*$ using source symbols (e.g., \&). Surprisingly, the model correctly outputs the object bound to the source symbol, even though no explicit occurrence of the symbol \& is present in the host context. The average prediction accuracy reaches $\mathbf{0.86}$, which is considerably higher than the random chance level.
Overall, these experiments uncover the nature of Grounding IDs from complementary perspectives (see also Appendix~\ref{sec:logit_lens} for complementary logit-lens analysis and proximity-controlled variants).

\section{Behavioral Implications of Grounding IDs} \label{sec:hallucination}
Our observations in Section \ref{sec:existence} exhibit a reduced modality gap and increased attention between related partitions in samples with external cues. The causal analysis in Section \ref{sec:causality} further reveals the presence of latent identifiers that bind to the embeddings of the corresponding partitions. %Together, these results describe a mechanism by which external cues improve visual reasoning: they slightly modify the model’s internal dynamics and equip visual embeddings with auxiliary identifiers that strengthen the model’s grounding ability. 
To examine how this property affects downstream behavior, we first evaluate the impact of Grounding IDs on hallucination mitigation during long caption generation, since this task directly depends on grounding. We then assess the effect of aligned multimodal cues on broader visual reasoning tasks.

\subsection{Hallucination Mitigation}

\begin{figure}[t]
  \centering
  \includegraphics[width=0.85\linewidth]{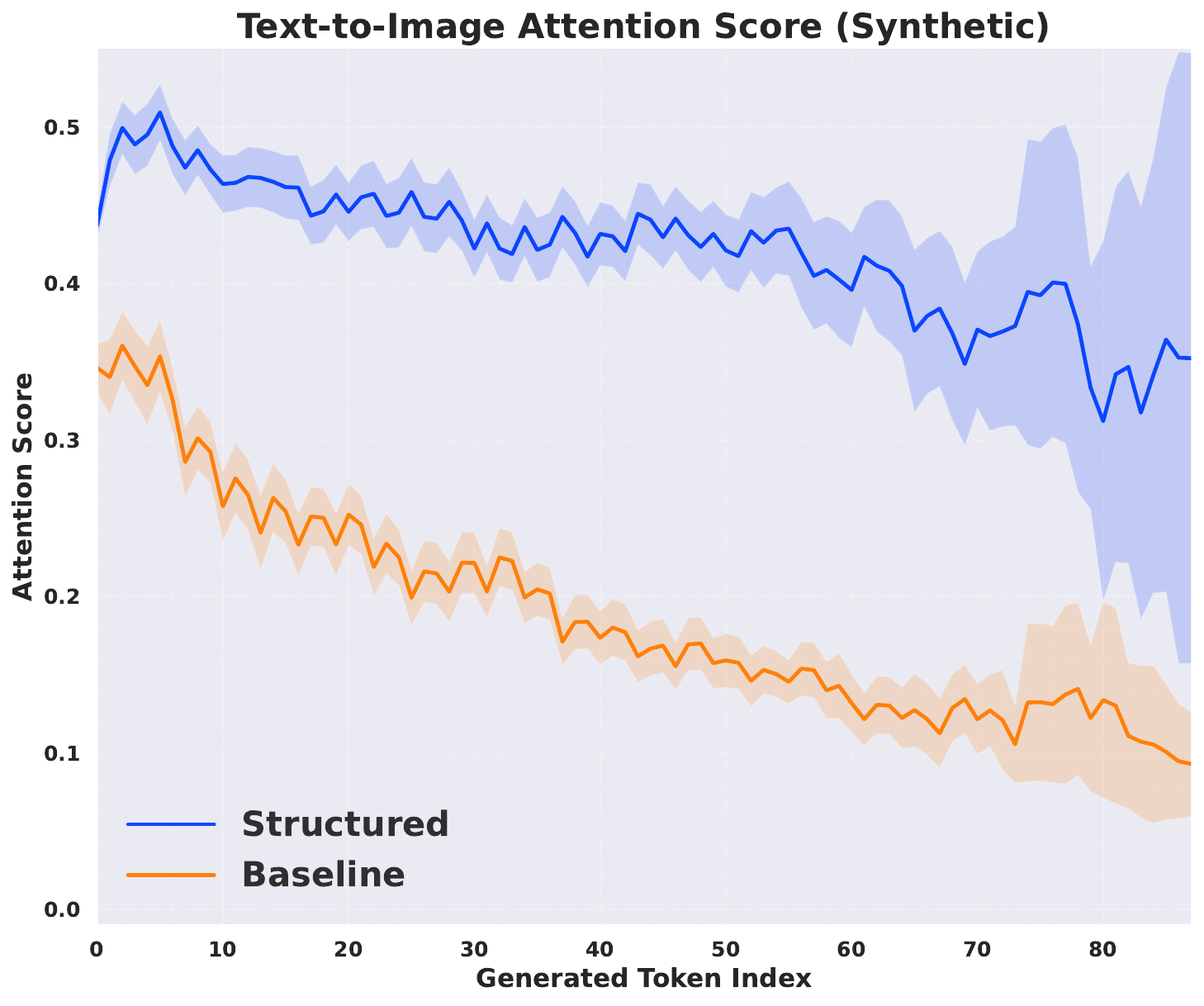}
  \vspace{-1mm}
    \caption{Averaged cross-attention behavior on the synthetic dataset with 20 objects.
    % averaged over 100 samples.
    Attention is computed using a max operation with a window size of 5 and stride of 1 across generated tokens.}
  \label{fig:attention_span_synthetic}
  \vspace{-1mm}
\end{figure}

In Section~\ref{sec:existence}, we reported overall attention enhancement across the bound partitions. To test whether this partition-based effect also helps LVLMs remain focused on the image in long responses, we measure cross-attention during generation.
Using the same synthetic dataset as in Section~\ref{sec:existence}, the model is asked to produce detailed descriptions of each image based on the provided external cues.
We compute cross-attention from generated tokens to image patches using a sliding window, taking the maximum value within each window of size 5.
Plotting these values against token position (Fig.~\ref{fig:attention_span_synthetic}) reveals a consistent decline: tokens appearing later in the response attend less to the image, indicating that visual grounding diminishes as generation progresses. 
Importantly, the structured case shows both higher initial attention and a slower rate of decline compared to the baseline. This indicates that external cues sustain grounding over longer spans of text.

\begin{table}[t]
  \centering
  \caption{Evaluation of the scene description task on synthetic datasets, each containing 500 samples with 10, 15, or
  20 unique objects per image.}
  \label{tab:synthetic_hallucination}
  \resizebox{\linewidth}{!}{%
    \begin{tabular}{l l c c c c}
    \toprule
    \textbf{\# Obj.} & \textbf{Method} & \textbf{Precision} & \textbf{Recall} & \textbf{F1} & \textbf{Acc.} \\
    \midrule

    \multirow{4}{*}{\textbf{10}} 
      & Baseline & 0.56 & 0.56 & 0.58 & 0.42 \\
      & {\small Structured (text-only)} & \underline{0.59} & \textbf{0.68} & \underline{0.63} & \underline{0.46} \\
      & {\small Structured (img-only)} & 0.53 & \underline{0.59} & 0.56 & 0.38 \\
      & \textbf{Structured (both)} & \textbf{0.74} & 0.58 & \textbf{0.65} & \textbf{0.48} \\
    \midrule

    \multirow{4}{*}{\textbf{15}} 
      & Baseline & 0.30 & 0.49 & 0.37 & 0.24 \\
      & {\small Structured (text-only)} & 0.33 & \textbf{0.61} & 0.44 & 0.27 \\
      & {\small Structured (img-only)} & \underline{0.43} & 0.51 & \underline{0.46} & \underline{0.30} \\
      & \textbf{Structured (both)} & \textbf{0.67} & \underline{0.53} & \textbf{0.59} & \textbf{0.46} \\
    \midrule

    \multirow{4}{*}{\textbf{20}} 
      & Baseline & 0.14 & 0.45 & 0.21 & 0.12 \\
      & {\small Structured (text-only)} & 0.29 & \underline{0.57} & 0.39 & \underline{0.24} \\
      & {\small Structured (img-only)} & \underline{0.39} & 0.42 & \underline{0.40} & \underline{0.24} \\
      & \textbf{Structured (both)} & \textbf{0.65} & \textbf{0.59} & \textbf{0.62} & \textbf{0.40} \\
      \bottomrule
    \end{tabular}%
  }
\end{table}

Prior work attributes hallucinations in LVLMs to the decay of visual attention and increasing reliance on language priors \cite{m3id, opera}. Motivated by this, we evaluate hallucination on the controlled datasets from Section~\ref{sec:existence},  as reported in Table~\ref{tab:synthetic_hallucination}.
Structured modalities consistently improve performance across all metrics, with particularly strong gains in precision, which is most relevant for avoiding mentions of nonexistent objects. For images with 10 objects, baseline recall is slightly higher, but as the number of objects increases, structured inputs outperform the baseline across all criteria. The performance gap due to external structure also widens as the number of objects increases. Notably, multimodal cues are considerably more effective than unimodal cues.
These results highlight that sustained cross-modal grounding directly improves faithfulness of generated text.

Finally, we evaluate on large-scale hallucination benchmarks using MS-COCO images \cite{ms_coco}, with performance assessed by the hallucination-specific metric CHAIR \cite{chair}. We conduct experiments on two recent LVLMs, Qwen2.5-VL \cite{qwen2.5_vl}, and LLaVA-1.5 \cite{llava1.5}. We observe that for natural scenes, simple horizontal lines are insufficient, and grid-based partitions provide more effective structure for sequential scanning of the image. To ensure visibility against diverse backgrounds, we add thin white margins around the scaffolding.

\begin{figure}[t]
  \centering
    \includegraphics[width=0.85\linewidth]{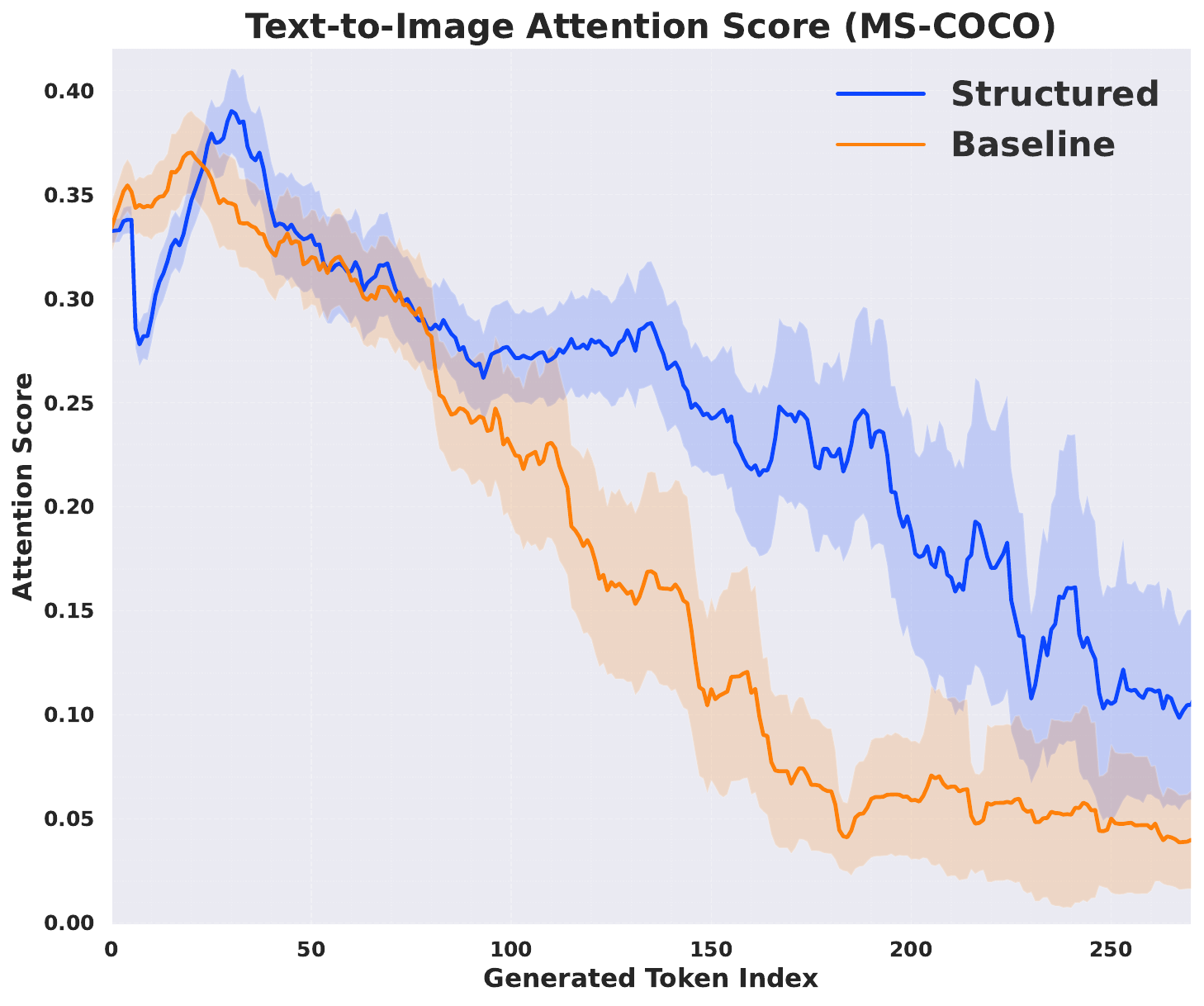}
    \vspace{-1mm}
    \caption{Cross-attention behavior on the MS-COCO dataset across generated tokens, computed using a max operation with a window size of 10 and stride of 1.}
    \label{fig:attention_span_coco}
    \vspace{-1mm}
\end{figure}

Results (Fig.~\ref{fig:attention_span_coco} and Table~\ref{tab:coco_hallucination}) show substantial reductions in both CHAIR$_s$ and CHAIR$_i$ compared to baseline. Importantly, unlike most existing techniques for hallucination mitigation, the proposed approach is also applicable to black-box, closed-source models such as GPT-4o \cite{gpt4o} and Gemini-2.5-Pro \cite{gemini2.5}, which are considered strong models for visual reasoning. Notably, our simple strategy outperforms or matches specialized hallucination-mitigation methods such as VCD \cite{vcd}, OPERA \cite{opera}, and SPARC \cite{sparc}, while requiring no additional inference modules and maintaining near-zero computational overhead.
% include some failure cases and
We also report results on additional hallucination benchmarks, including POPE \cite{pope} and ROPE \cite{rope}, in Appendices~\ref{appendix:pope_results} and~\ref{appendix:rope_results}.

\begin{table}[t]
  \centering
    \centering
    \captionof{table}{Evaluation on 500 MS-COCO samples using CHAIR metrics across open- and closed-source models. Results are reported for sentence-level (CHAIR$_s$) and instance-level (CHAIR$_i$) hallucination rates.
    % The average inference time measured on a machine with an NVIDIA RTX 4090 GPU.
    }
    \label{tab:coco_hallucination}
    \resizebox{\linewidth}{!}{%
    \begin{tabular}{l l  c c c}
    \toprule
    \textbf{Model} & \textbf{Method}  & \textbf{CHAIR$_s$$\downarrow$} & \textbf{CHAIR$_i$$\downarrow$} & \textbf{{\small Inf. Time(s)}} \\
    \midrule
    \multirow{5}{*}{\textbf{LLaVA-1.5}}
    & Baseline         & \underline{51.60}  & 13.20 & 3.41 \\
    & OPERA        & 48.00  & 13.52 & 20.91\\         
    & VCD          & 54.40  & 14.28 & 7.81\\
    & SPARC        & 55.20  & \underline{12.78} & 4.50\\
    & \textbf{Structured}         & \textbf{41.00} & \textbf{12.04} & 3.94\\
    \midrule
    \multirow{5}{*}{\textbf{Qwen2.5-VL}}
    & Baseline         & 32.40  & \underline{7.97} & 3.31\\
    & OPERA        & \underline{29.60} & 10.76 & 23.50\\         
    & VCD          & 33.80  & 8.91 & 9.73\\
    & SPARC        & 33.60  & 8.21 & 5.50\\
    & \textbf{Structured}         & \textbf{27.20} & \textbf{5.36} & 6.04\\
    \midrule
    \multirow{2}{*}{\textbf{GPT-4o}}
    & Baseline         & \underline{29.20} & \underline{6.40} & -\\
    & \textbf{Structured}         & \textbf{23.20} & \textbf{5.81} & -\\
    \midrule
    \multirow{2}{*}{\textbf{Gemini-2.5-Pro}}
    & Baseline         & \underline{44.20} & \underline{8.64} & -\\
    & \textbf{Structured}         & \textbf{37.40} & \textbf{7.28} & -\\
    \bottomrule
    \end{tabular}%
  }
\end{table}

\subsection{Visual Reasoning Performance}

We evaluate the effectiveness of modality-aligned cues on two visual reasoning benchmarks, using the same experimental setup as \citet{viser}, and compare it with the existing method, VISER. The results for the counting and visual search tasks are shown in Table~\ref{tab:reasoning}. 

We evaluate three models: Qwen2.5-VL (3B), Qwen2.5-VL (7B), and GPT-4o. Grounding IDs increase accuracy on both tasks and outperform the VISER baseline and the unstructured input. The gains depend on two conditions: each partition must have a distinct identifier, and the same identifiers must appear in the image and the prompt.

\begin{table}[H]
  \centering
  \vspace{-2mm}  % Reduce the gap before the table
  \caption{Performance on counting and visual search benchmarks.}
  \label{tab:reasoning}
  \vspace{-3mm}  % Adjust space before the table
  \resizebox{\linewidth}{!}{%
    \begin{tabular}{lccc}
    \toprule
    \multicolumn{4}{c}{\textbf{Counting Accuracy}} \\
    \midrule
    \textbf{Model} & \textbf{Baseline} & \textbf{VISER} & \textbf{Grounding IDs} \\
    \midrule
    \textbf{Qwen2.5-VL (3B)} & 30.00 & \underline{37.83} & \textbf{43.00} \\
    \textbf{Qwen2.5-VL (7B)} & 29.67 & \underline{43.33} & \textbf{53.00} \\
    \textbf{GPT-4o}         & 10.50 & \underline{26.50} & \textbf{32.33} \\
    \midrule
    \multicolumn{4}{c}{\textbf{Visual Search Accuracy}} \\
    \midrule
    \textbf{Model} & \textbf{Baseline} & \textbf{VISER} & \textbf{Grounding IDs} \\
    \midrule
    \textbf{Qwen2.5-VL (3B)} & 0.00 & \underline{37.83} & \textbf{45.96} \\
    \textbf{Qwen2.5-VL (7B)} & 30.00 & \underline{40.00} & \textbf{52.25} \\
    \textbf{GPT-4o}         & 49.41 & \underline{73.40} & \textbf{80.62} \\
    \bottomrule
    \end{tabular}
  }
  \vspace{-4mm}  % Reduce space after the table
\end{table}

\section{Related Work}

\paragraph{Interpretability.}
Interpretability research investigates where and how information is represented and routed in neural networks. For \textbf{LLMs}, attention analyses study which tokens heads attend to and how this shapes predictions~\cite{chefer21_generic_attention,clark19_bert_attention}, while circuit-oriented methods identify causal substructures and enable editing through activation or circuit discovery and factual localization~\cite{conmy23_automated_circuit_discovery,meng22_rome_factual_edit,geva23_dissect_recall}. Probing tools such as the \emph{logit lens} decode intermediate states to reveal how token predictions evolve across layers~\cite{nostalgebraist20_logit_lens}.
For \textbf{VLMs}, recent work explores where visual information is stored and how it is transferred into the language pathway~\cite{basu24_storage_transfer_vlm}, and adapts logit-lens–style spatial probes to study grounding and localization inside multimodal models~\cite{jiang24_vl_rep_edit,llava_interp}.
% In parallel, analyses of the \emph{modality gap} in contrastive VLMs show that a small set of embedding dimensions drives the gap and that \emph{information imbalance} between images and captions is a key trigger~\cite{modality_gap}.
Overall, the interpretability methods provide the toolset we build on: causal interventions and layerwise probes to expose mechanisms underlying cross-modal binding and grounding.

\paragraph{Abstract latent variables in representation space.}
Recent advances suggest that models internally rely on abstract latent variables to maintain entity–attribute associations. In LLMs, \emph{Binding IDs} are content-independent identifiers that link entities and attributes through a shared latent code; causal interventions that swap these vectors systematically change the inferred associations~\cite{binding_id}. Follow-up work localizes a low-rank subspace that \emph{causally} governs which entity pairs with which attribute~\cite{dai24_binding_order}.
In multimodal models, analogous identifiers appear in VLMs: distinct binding codes are attached to an object’s image tokens and its textual mentions, yielding cross-modal referential consistency~\cite{vision_binding_id}. Complementary evidence points to \emph{symbolic indexing} in VLMs, where attention heads compute content-independent spatial indices and later retrieve attributes by these indices, thereby implementing object-centric binding across modalities~\cite{visual_symbolic_mechanism}.
% Related learning dynamics in synthetic program-tracking tasks show that Transformers acquire variable binding by using the residual stream as an addressable memory and specialized heads to route bindings across positions~\cite{wu25_var_binding}.
These findings highlight the role of hidden symbolic variables in reasoning. While prior analyses primarily examine existing identifiers, we show that simple input-augmented artifacts can \emph{elicit} partition-specific identifiers that are linked across modalities and enhance grounding through tighter cross-modal attention and alignment.

\paragraph{LVLM reasoning.}
% VoT
Efforts to improve reasoning in LVLMs have followed two main directions. Prompt-focused methods such as chain-of-thought prompting, majority voting, and test-time compute scaling aim to enhance reasoning without altering inputs. However, evaluations on multimodal reasoning benchmarks show that these approaches remain ineffective for tasks requiring visual or spatial reasoning, even when scaled substantially \cite{emma}. In contrast, recent studies demonstrate that external scaffolding (i.e., adding annotations, grids, or partitions to the input image) consistently improves model performance across counting, spatial relations, and description tasks \cite{forgotten_polygons,viser}. While such findings establish the effectiveness of scaffolding, the mechanisms underlying these improvements remain poorly understood.

\paragraph{Hallucination reduction.}
Hallucination remains a central challenge for LVLMs, mainly arising from the decline of cross-modal attention during long text generation and over-reliance on language priors.
Several inference-time methods have been proposed, including VCD \cite{vcd}, which down-weights tokens favored under distorted inputs; OPERA \cite{opera}, which applies an over-trust penalty and retrospection allocation to rebalance attention; and SPARC \cite{sparc}, which progressively recalibrates visual attention to maintain relevance in long captions. Our work complements these approaches by identifying how visual scaffolding induces Grounding IDs that naturally sustain cross-modal binding, thereby reducing hallucination without additional inference tools.

% \section{Discussion and Future Work}
\section{Conclusion}

In this work, we introduce a conceptual framework to explain how multimodal aligned external cues induce abstract \textit{Grounding IDs} across related partitions. Through attention and embedding alignment analysis, we showed that these identifiers propagate across modalities, supporting partition-specific grounding and reducing the modality gap. Activation patching interventions confirmed that the associations are mediated by abstract identifiers rather than local features. Empirical evaluations further demonstrated that Grounding IDs enhance cross-modal attention, yield more faithful image descriptions, reduce hallucinations, and improve visual reasoning. Since the approach relies only on simple, content-independent structures, it provides a model-agnostic strategy applicable to a wide range of tasks and models, including closed-source LVLMs.

Beyond empirical contributions, our methodology provides insights into mechanistic interpretability in multimodal models through both observational and causal tools. In addition, this study opens several directions for future research. One direction involves discovering circuits that support system-2 reasoning tasks, such as counting and spatial reasoning. Another arises from a key observation in this study: external cues in VLM inputs can reinforce the model's inherent grounding capability. Building on this, future work could explore integrating such cues during RL fine-tuning to further strengthen the model’s internal ability to perform sequential scanning based on provided cues.

\section*{Impact Statement}

This work aims to advance the understanding and interpretability of vision–language models by studying how simple, aligned external cues improve multimodal grounding and reduce hallucinations. The techniques explored are model-agnostic and rely on lightweight input modifications, which may contribute to safer and more reliable deployment of multimodal systems in applications that require faithful visual grounding. We do not anticipate significant negative societal impacts beyond those already associated with large-scale vision–language models, and this work does not introduce new data sources, learning objectives, or deployment scenarios that raise additional ethical concerns.

\clearpage

\bibliography{bib}
\bibliographystyle{icml2026}

\clearpage
\appendix
\onecolumn
% \section{Appendix}

\section{Visual Traversal and Positional Inductive Bias}
\label{sec:att_1}
To analyze how the model traverses the visual input, we compute mean
cross-attention maps from textual tokens to image rows. These maps are averaged
across 100 samples, respectively, and aggregated over layers 14--28. Each
column corresponds to a decoding step, while rows are grouped top-to-bottom.
This setup allows us to observe how attention flows across the image during
generation.

\begin{figure}[ht]
    \centering
    \begin{subfigure}{0.9\textwidth}
        \centering
        \includegraphics[width=\linewidth]{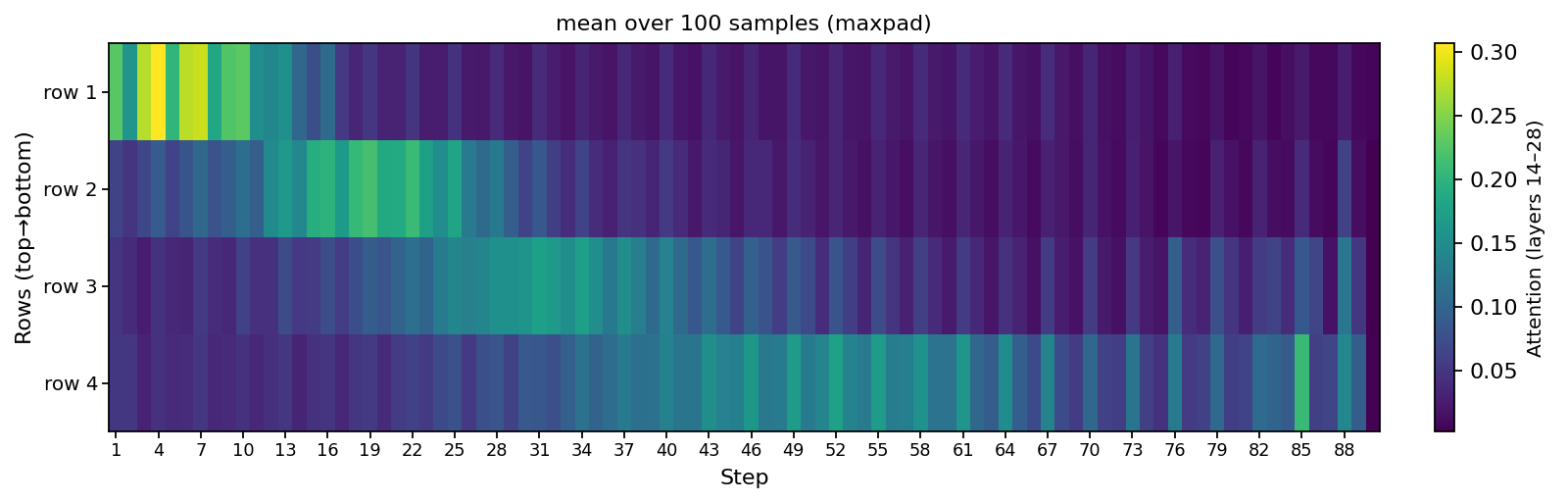}
        \caption{Baseline input without structural cues (Averaged over 100 samples).}
        \label{fig:heatmap_base}
    \end{subfigure}

    \vspace{0.5em}

    \begin{subfigure}{0.9\textwidth}
        \centering
        \includegraphics[width=\linewidth]{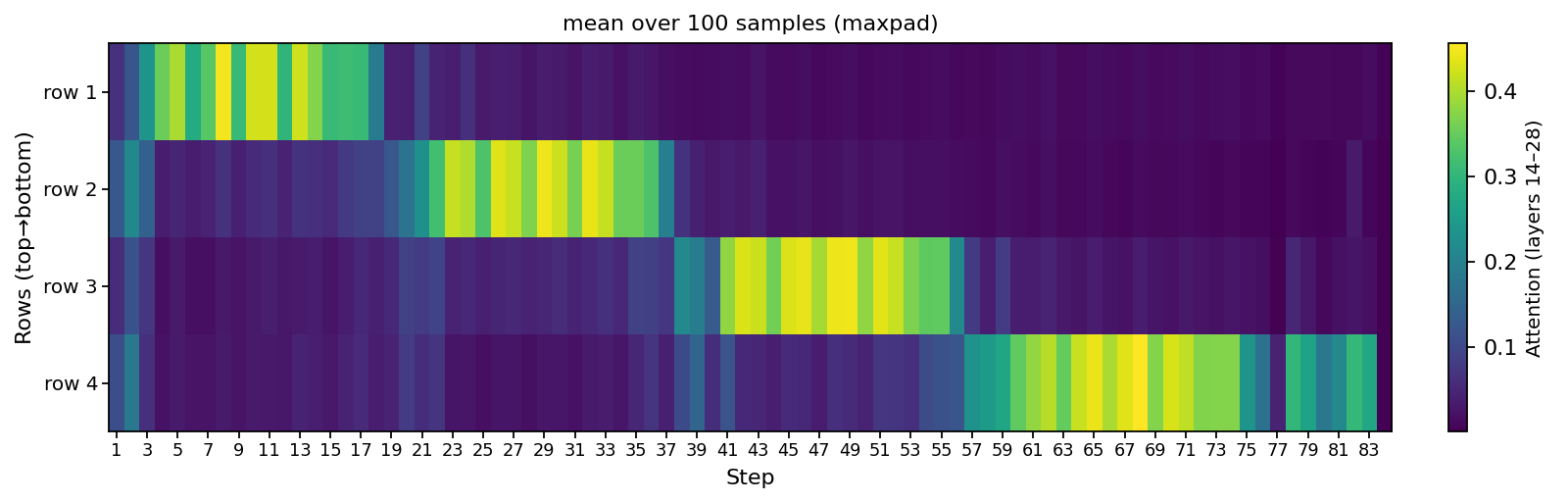}
        \caption{
        Structured input with explicit row cues (Averaged over 100 samples).
        }
        \label{fig:heatmap_structured}
    \end{subfigure}

    \caption{Mean cross-attention maps (layers 14–28) from textual tokens to image rows. Structural cues amplify the model’s inherent top-to-bottom, left-to-right traversal bias and produce sharper row-aligned attention patterns.}
    \label{fig:heatmap_comparison}
\end{figure}

Fig.~\ref{fig:heatmap_comparison} reveals two
consistent patterns:

\begin{enumerate}
    \item \textbf{Inherent top-to-bottom, left-to-right traversal.}
    Even without explicit guidance, the model exhibits a natural reading-like
    behavior: starting from the upper-left region of the image and progressively
    shifting attention downward. This reflects the \emph{positional encoding inductive bias} of the vision tokenizer, which encourages sequential exploration and may facilitate grounding by providing a consistent spatial reference across tokens.

    \item \textbf{Effect of structural cues.}
    When explicit structure is added to the image (e.g., row separators or
    symbolic markers), the attention becomes sharper and more aligned with the
    intended reading order. Instead of diffuse activation, the model follows a
    clear row-by-row progression from top to bottom. This demonstrates how structural cues can \emph{reinforce inherent grounding}, leading to more consistent visual traversal.   
\end{enumerate}

\begin{figure}[ht]
    \centering
    \includegraphics[width=0.95\linewidth]{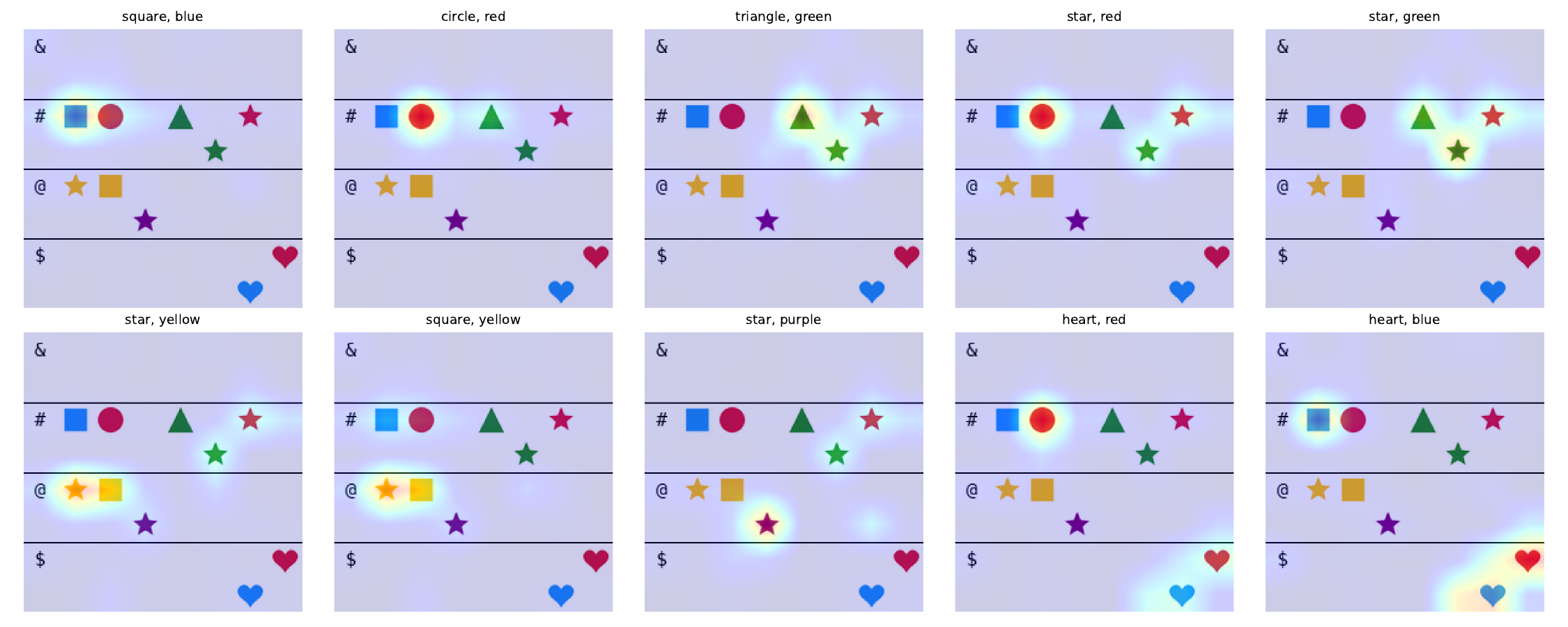}
    \caption{Examples of per-object traversal heatmaps for a structured image. Each panel shows the attention distribution for a specific shape–color pair. Structural cues guide the model toward sharper localization and promote systematic
    top-to-bottom and left-to-right traversal.}
    \label{fig:object_traversal}
\end{figure}

As a qualitative example, Fig.~\ref{fig:object_traversal} illustrates cross-attention heatmaps for individual objects. Each panel corresponds to a specific shape–color pair, showing how the model localizes and attends to the correct region during decoding. This highlights the row-by-row traversal pattern on a per-object basis.

\begin{figure}[ht]
    \centering
    \includegraphics[width=0.95\linewidth]{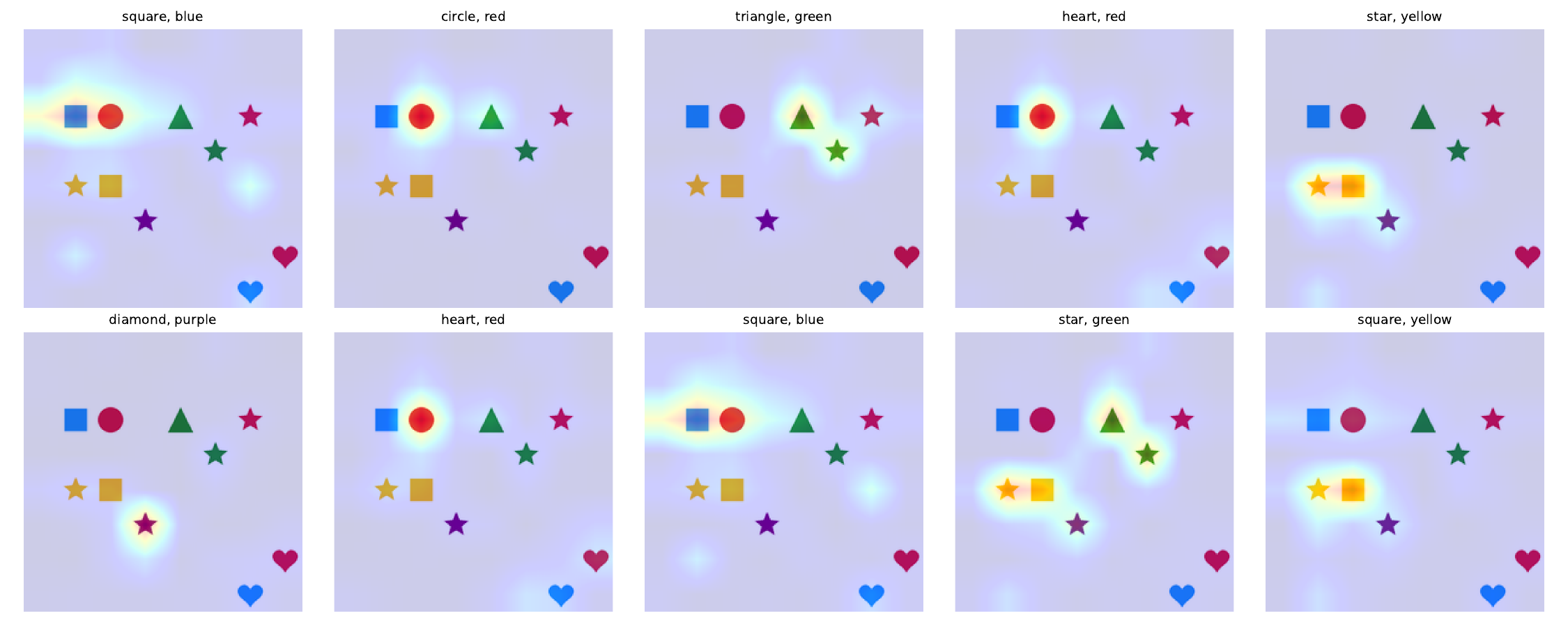}
    \caption{Examples of per-object traversal heatmaps for the baseline. Each panel shows the attention distribution for a specific shape–color pair. The baseline also exhibits a general top-to-bottom and left-to-right pattern, but accuracy degrades over generation.}
    \label{fig:based_traversal}
\end{figure}

Overall, these results indicate that the model does not scan the visual field randomly. Instead, it exhibits a systematic traversal strategy that can be strengthened by structural modifications to the input image, yielding more interpretable and faithful cross-modal grounding.

\section{Intra-Visual Attention Patterns}
\label{sec:att_2}
A key signal of grounding is whether patches from the same row interact more strongly, indicating that the model encodes rows as coherent positional units rather than arbitrary patch collections.
To probe this effect, we compute pairwise cosine similarities between final-layer patch embeddings and compare the average similarity of patches within the same row to that of patches drawn from different rows. A clear gap between within-row and across-row similarity demonstrates that structural cues encourage row-wise grouping, effectively injecting an implicit positional prior into the representation space.

\begin{figure}[H]
    \centering
    \includegraphics[width=0.50\linewidth]{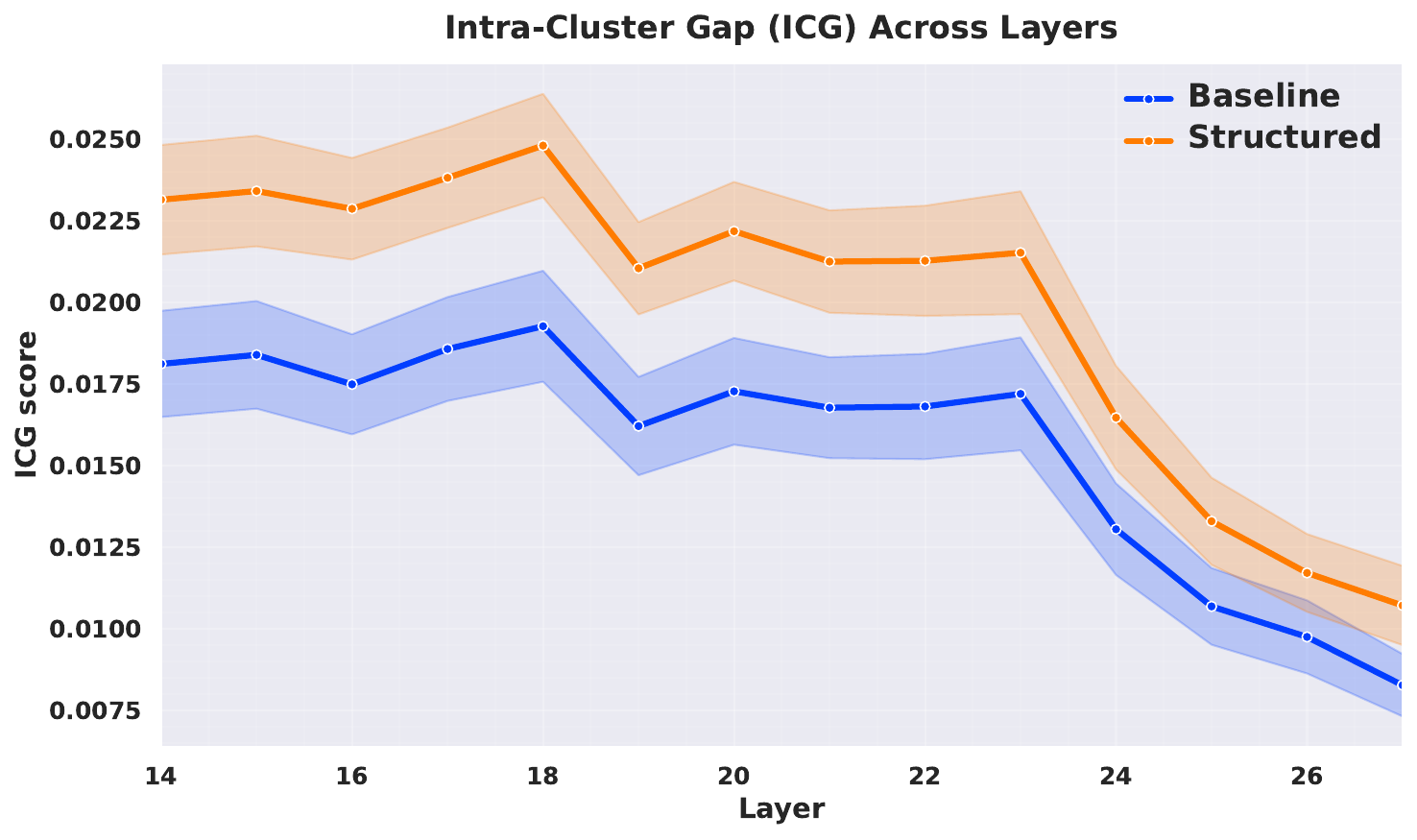}
    \vspace{-2mm}
    \caption{ICG scores across layers for baseline vs. structured inputs. Structured inputs consistently yield higher ICG, reflecting stronger row-level grouping.}
    \label{fig:icg_layers}
    \vspace{-2mm}
\end{figure}

We further quantify this phenomenon with the \textbf{intra--inter cluster
grounding (ICG)} score, defined as the difference between the mean within-row
similarity and the mean across-row similarity:
\begin{equation}
\mathrm{ICG} =
\frac{1}{R}\sum_{r=1}^{R} \text{Sim}_{\mathrm{intra}}(r)
\;-\;
\frac{1}{R(R-1)} \sum_{r \neq r'} \text{Sim}_{\mathrm{inter}}(r,r'),
\label{eq:icg}
\end{equation}
where $R$ is the number of rows, $\text{Sim}_{\mathrm{intra}}(r)$ is the mean
cosine similarity between patches within row $r$, and
$\text{Sim}_{\mathrm{inter}}(r,r')$ is the average similarity between patches
across rows $r$ and $r'$. A larger $\mathrm{ICG}$ score indicates stronger
row-wise clustering and hence a clearer positional alignment in the image
representation.
Fig.~\ref{fig:icg_layers} plots the ICG score across layers for both the
baseline and the structured setting. We observe that the structured condition
(symbol-augmented inputs) consistently achieves higher ICG values than the
baseline, indicating that structural cues enhance row-level grouping.

\section{Logit Lens Analysis and Evidence for Propagation}
\label{sec:logit_lens}

To better understand how symbolic cues influence within-partition binding, we apply a \emph{logit lens} analysis to structurally manipulated images from our synthetic dataset containing seven distinct shapes and five colors. We project patch hidden states into the vocabulary space via the unembedding matrix, applying a softmax over the four symbol tokens (\&, \#, @, \$). This measures how strongly each visual patch activates each symbol, revealing how symbolic evidence propagates through the visual representation.

\begin{figure}[H]
    \centering
    \includegraphics[width=0.85\linewidth]{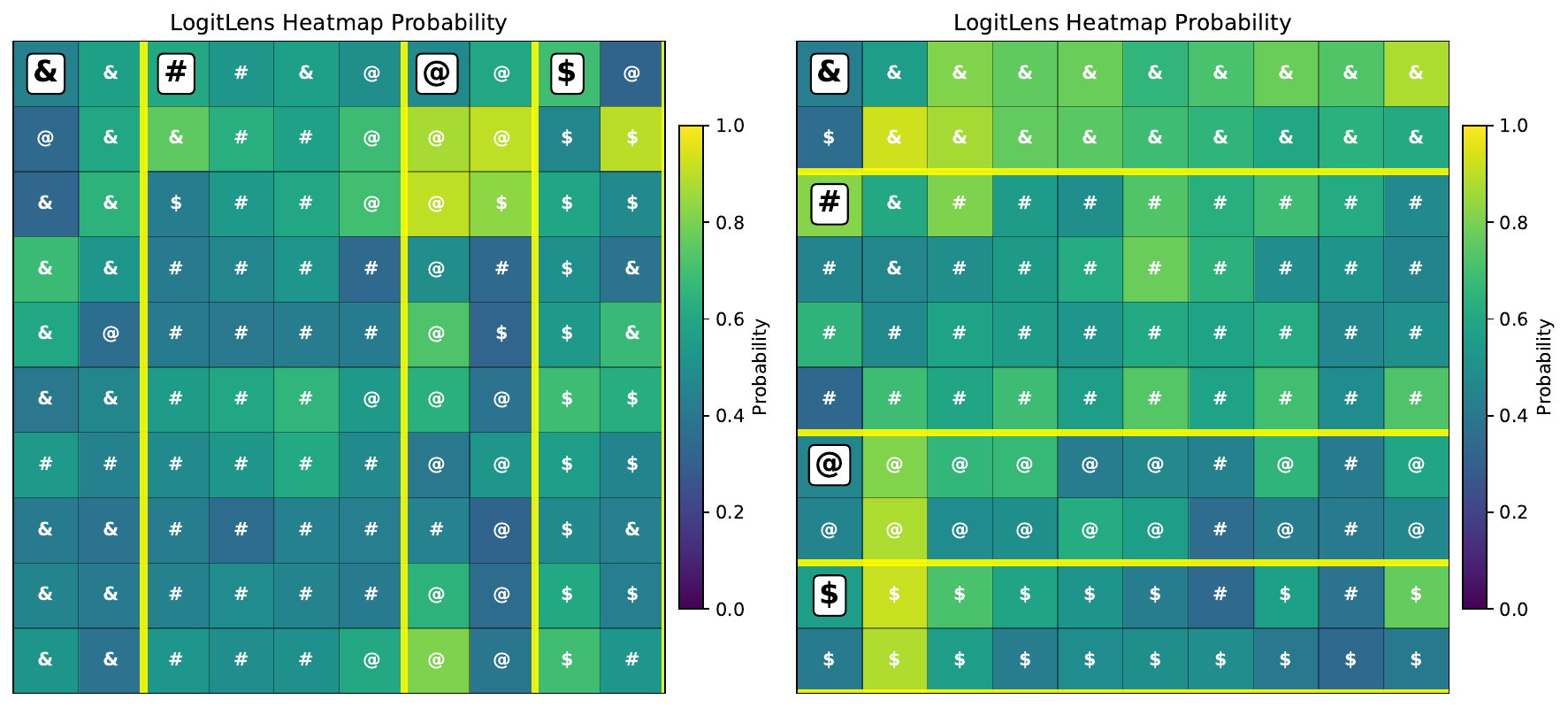}
    \vspace{-2mm}
    \caption{
    Proximity-controlled logit-lens analysis. The two panels test whether symbol propagation can be explained only by local adjacency to the nearest symbol. In both layouts, visual tokens are mainly associated with their partition symbol rather than simply the closest symbol, arguing against a purely local semantic-blurring explanation.
    }
    \label{fig:logit_lens}
    \vspace{-2mm}
\end{figure}

A possible alternative explanation is that object patches simply absorb the nearest symbolic context through local late-layer attention, rather than carrying a structured partition-level identifier. To test this, we use proximity-controlled layouts that break the alignment between partition membership and nearest-symbol proximity. In one layout, the spatial structure disrupts the symmetry between partitions and local distance to symbols; in another, the partitioning is less aligned with the model's natural top-to-bottom traversal. As shown in Fig.~\ref{fig:logit_lens}, visual tokens are still primarily associated with their partition symbol rather than merely with the closest symbol. This indicates that the observed propagation is not reducible to local semantic blurring around nearby symbols, and instead supports structured partition-level binding induced by the external cues.

\section{Evaluation on Synthetic Variants}
\label{sec:cue:ablation}

We evaluate a range of structural variants of the synthetic dataset 
(7 shapes × 5 colors, with 20 objects per image) to understand how explicit
row-level cues affect grounding. Each row in 
Table~\ref{tab:metrics_variants} reports \textbf{precision}, 
\textbf{recall}, \textbf{F1}, and \textbf{accuracy}. 

The baseline condition provides no additional cues and yields
weak performance. Introducing explicit structure consistently improves results,
with the following variants:

\begin{itemize}
    \item \textbf{Numbers (no line)}: Each row is tagged with a numeric index,
    but no horizontal separators are drawn. This already improves both precision
    and accuracy compared to the baseline.
    
    \item \textbf{Letters + Line}: Rows are separated by horizontal lines and
    tagged with distinct letter labels. This variant yields a significant boost
    in precision, indicating clearer disambiguation between rows.
    
    \item \textbf{Numbers + Line}: Rows are separated by lines and also numbered.
    This combination achieves the highest recall among all variants, suggesting
    that explicit row boundaries help the model correctly cover more objects.
    
    \item \textbf{Symbols + Line}: Each row is separated by a line and prefixed
    with a unique symbol (\&, \#, @, \$). This provides the strongest positional
    anchor, achieving the best overall F1 and accuracy, reflecting both balanced
    precision and recall.
    
    \item \textbf{Grid 4×4}: A baseline layout where objects are arranged in a $4 \times 4$ grid. Two variants were tested: numbered with grid lines, and numbered without grid lines. Performance remained weaker than in the structured row version.

    \item \textbf{Numbers Right-Aligned + Line}: A variant of
    \texttt{Numbers + Line} where the numeric label appears at the end of the
    row rather than the beginning. While recall improves, the overall precision
    drops slightly, indicating less consistent alignment.

    \item \textbf{Symbols (no line)}: Each row is tagged with a unique symbol (\&, \#, @, \$), but no horizontal separators are drawn. This already improves both precision and accuracy compared to the baseline.

    \item \textbf{Object Bounding (input-aware)}: Each object is enclosed within an explicit bounding box. While this provides a localized spatial cue, it is less effective than row-level structuring, resulting in only moderate improvements over the
   baseline.

   \item \textbf{Symbols on Objects (input-aware)}: Symbols are assigned at the object level
instead of the row level. While this introduces additional visual markers, the
lack of global row structure leads to reduced performance compared to
row-based variants.
\end{itemize}

% \begin{table}[t]
% \centering
% \setlength{\tabcolsep}{8pt}
% \renewcommand{\arraystretch}{1.2}
% \resizebox{\textwidth}{!}{%
% \begin{tabular}{lcccc}
% \toprule
% \textbf{Variant} & \textbf{Precision} & \textbf{Recall} & \textbf{F1} & \textbf{Accuracy} \\
% \midrule
% Baseline       & 0.14 & 0.45 & 0.21 & 0.12 \\
% Numbers (no line)       & 0.53 & 0.43 & 0.47 & 0.31 \\
% Letters + Line          & \underline{0.64} & 0.48 & 0.55 & \underline{0.38} \\
% Numbers + Line          & 0.56 & \underline{0.58} & \underline{0.57} & \textbf{0.40} \\
% Symbols + Line          & \textbf{0.65} & \textbf{0.59} & \textbf{0.62} & \textbf{0.40} \\
% Grid 4×4 Numbered       & 0.33 & 0.40 & 0.36 & 0.22 \\
% Grid 4×4 Numbered (no lines) & 0.33 & 0.35 & 0.34 & 0.20 \\
% Numbers Right-Aligned + Line & 0.39 & 0.55 & 0.46 & 0.30 \\
% Symbols (no line)       & 0.48 & 0.50 & 0.49 & 0.33 \\
% Object Bounding Box       & 0.43 & 0.48 & 0.44 & 0.32 \\
% Symbols on Objects       & 0.30 & 0.38 & 0.29 & 0.19 \\
% \bottomrule
% \end{tabular}%
% }
% \caption{Performance metrics (precision, recall, F1, accuracy) on the scene description task across synthetic variants with 20 objects. Explicit structural cues (such as lines, labels, or symbols) consistently improve performance compared to the baseline.}
% \label{tab:metrics_variants}
% \end{table}

\begin{table}[t]
\centering
\caption{Performance metrics (precision, recall, F1, accuracy) on the scene description task across synthetic variants with 20 objects. Explicit structural cues (such as lines, labels, or symbols) consistently improve performance compared to the baseline.}
\setlength{\tabcolsep}{8pt}
\renewcommand{\arraystretch}{1.2}
\begin{tabular}{lcccc}
\toprule
\textbf{Variant} & \textbf{Precision} & \textbf{Recall} & \textbf{F1} & \textbf{Accuracy} \\
\midrule
Baseline       & 0.14 & 0.45 & 0.21 & 0.12 \\
Numbers (no line)       & 0.53 & 0.43 & 0.47 & 0.31 \\
Letters + Line          & \underline{0.64} & 0.48 & 0.55 & \underline{0.38} \\
Numbers + Line          & 0.56 & \underline{0.58} & \underline{0.57} & \textbf{0.40} \\
Symbols + Line          & \textbf{0.65} & \textbf{0.59} & \textbf{0.62} & \textbf{0.40} \\
Grid 4$\times$4 Numbered       & 0.33 & 0.40 & 0.36 & 0.22 \\
Grid 4$\times$4 Numbered (no lines) & 0.33 & 0.35 & 0.34 & 0.20 \\
Numbers Right-Aligned + Line & 0.39 & 0.55 & 0.46 & 0.30 \\
Symbols (no line)       & 0.48 & 0.50 & 0.49 & 0.33 \\
Object Bounding Box       & 0.43 & 0.48 & 0.44 & 0.32 \\
Symbols on Objects       & 0.30 & 0.38 & 0.29 & 0.19 \\
\bottomrule
\end{tabular}%
\label{tab:metrics_variants}
\end{table}

\FloatBarrier
\section{Implementation Details and Qualitative Example}

We employed two types of prompts for instruction tuning and evaluation:
a \emph{structured prompt} that explicitly encodes row-level layout with
symbols, and a simpler \emph{baseline prompt} without structural cues.

\paragraph{Structured Prompt.}
\begin{quote}
\textbf{Task:} You are presented with an image containing multiple objects.  
Your task is to identify all objects with color in the image.  
Each object must be exactly \texttt{<color> <shape>}.  
Scan the image sequentially based on horizontal lines and rows that exist in the image.  
There are exactly 4 horizontal rows in the image.  

Use \emph{only} these shapes (exact strings):  
\texttt{square, circle, triangle, diamond, star, moon, heart}  

Use \emph{only} these colors (exact strings):  
\texttt{red, green, blue, yellow, purple}  

Important: If the same object appears multiple times in a row, repeat its label
for each instance (e.g., \texttt{blue square, blue square, red circle}).  

\textbf{Output format (print \emph{exactly} this structure, no extra text):}  
\begin{verbatim}
Row &: <label>, <label>, ...
Row #: <label>, <label>, ...
Row @: <label>, <label>, ...
Row $: <label>, <label>, ...
\end{verbatim}
\end{quote}

\paragraph{Baseline Prompt.}
\begin{quote}
\textbf{Task:} You are presented with an image containing multiple colored objects.  
Your task is to identify all objects in the image and report each as \texttt{<color> <shape>}.  

Use \emph{only} these shapes (exact strings):  
\texttt{square, circle, triangle, diamond, star, moon, heart}  

Use \emph{only} these colors (exact strings):  
\texttt{red, green, blue, yellow, purple}  

If the same object appears multiple times in a row, repeat its label
for each instance (e.g., \texttt{blue square, blue square, red circle}).  

% \textbf{Output format (print \emph{exactly} this structure, no extra text):}  
% \begin{verbatim}
% Output a single line: a comma-and-space separated list of labels,
% where each label is "<color> <shape>".
% \end{verbatim}

\noindent\texttt{Output a single line: a comma-and-space separated list of labels,}\\
\texttt{where each label is "<color> <shape>".}

\end{quote}

% \paragraph{Qualitative Comparison.}
% \begin{figure}[ht]
%     \centering
%     % Structured image
%     \begin{subfigure}{0.41\textwidth}
%         \centering
%         \fbox{\includegraphics[width=\linewidth]{figs_appendix/image_045.png}}
%         \caption{Structured input with row cues.}
%         \label{fig:structured_input}
%     \end{subfigure}
%     \hfill
%     % Baseline image
%     \begin{subfigure}{0.41\textwidth}
%         \centering
%         \fbox{\includegraphics[width=\linewidth]{figs_appendix/base.png}}
%         \caption{Baseline input without cues.}
%         \label{fig:baseline_input}
%     \end{subfigure}
%     \caption{Comparison between structured and baseline prompts for the same image.}
%     \label{fig:structured_vs_base}
% \end{figure}

\noindent \textbf{Structured Prompt Output:}
% \begin{verbatim}
% Row &: purple moon, purple star, blue moon
% Row #: yellow heart, blue star, purple diamond
% Row @: blue circle, purple circle
% Row $: yellow triangle, blue heart
% \end{verbatim}
\noindent\texttt{Row \&: purple moon, purple star, blue moon}\\
\texttt{Row \#: yellow heart, blue star, purple diamond}\\
\texttt{Row \@: blue circle, purple circle}\\
\texttt{Row \$: yellow triangle, blue heart}

\noindent \textbf{Baseline Prompt Output:}
% \begin{verbatim}
% purple moon, purple star, purple diamond, yellow heart,
% blue circle, blue square, blue circle, blue circle,
% blue circle, blue circle, yellow triangle, blue heart
% \end{verbatim}
\noindent\texttt{purple moon, purple star, purple diamond, yellow heart,}\\
\texttt{blue circle, blue square, blue circle, blue circle,}\\
\texttt{blue circle, blue circle, yellow triangle, blue heart}

% \section{Qualitative Examples of our method}
% \label{appendix:qualitative_examples}

% To complement our quantitative results, we present a qualitative example that illustrates how structured scaffolding improves visual grounding and reduces hallucination in LVLM outputs.
% We show side-by-side captions with and without scaffolding, highlighting hallucinated objects (in red) in the baseline output.

\begin{figure*}[h]
    \centering
    \includegraphics[width=0.8\linewidth]{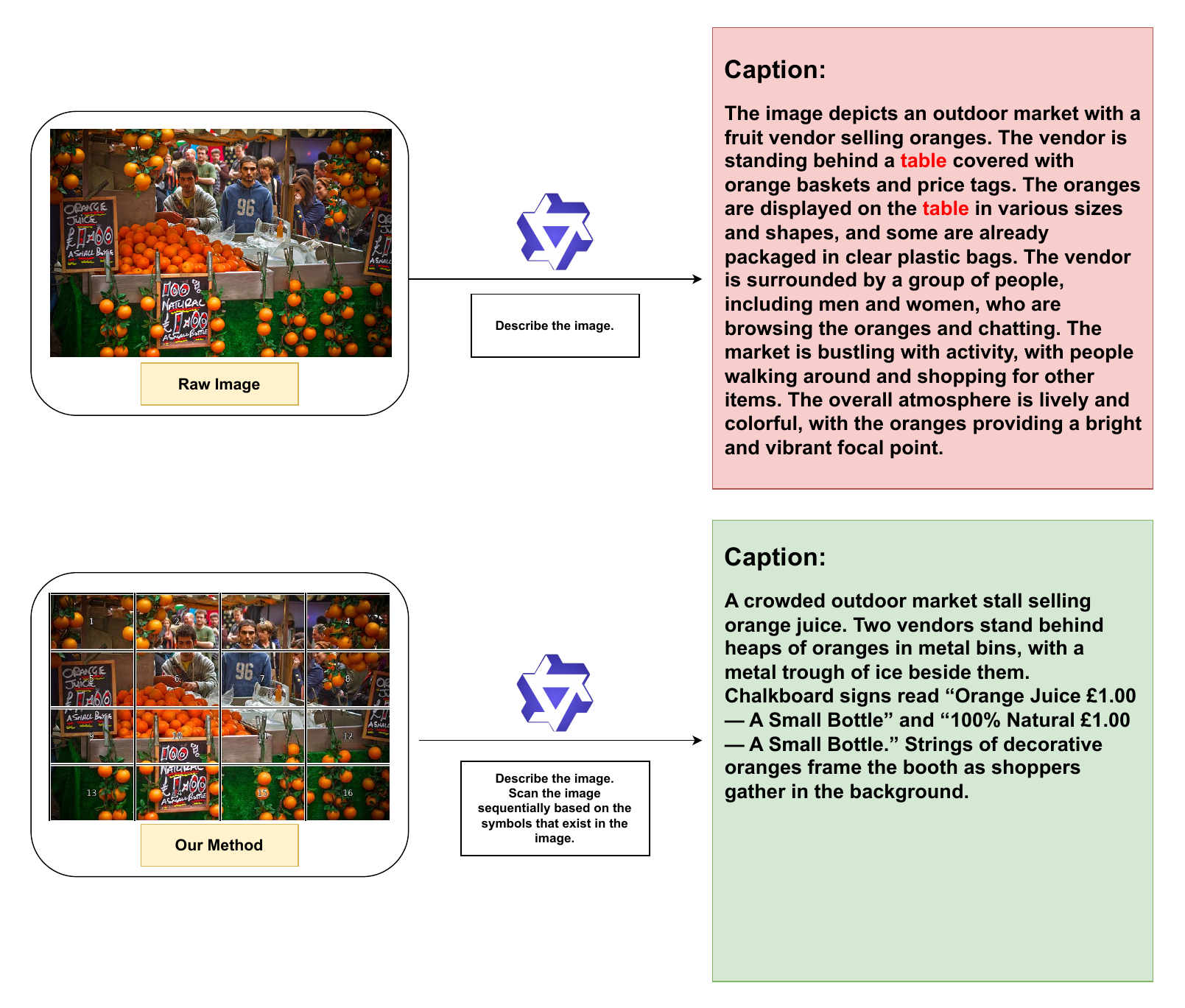}
    \caption{Qualitative example comparing captions generated with and without structured scaffolding. The figure shows the input image, the applied scaffolding structure, the caption produced by our method, and the corresponding baseline caption. Hallucinated objects in the baseline output are highlighted in \textcolor{red}{red}.}
\end{figure*}

\section{Evaluation Benchmarks for Object Hallucination}
\label{appendix:hallucination_metrics}

Object hallucination is a key failure mode of LVLMs, occurring when models generate content that is unsupported by the visual input. To quantify this phenomenon, we adopt two complementary evaluation frameworks: \textbf{CHAIR} and \textbf{POPE}. While both focus on grounding and image relevance, they differ in format, granularity, and the types of hallucination they expose.

\section{CHAIR: Caption Hallucination Assessment with Image Relevance}

The \textit{CHAIR} metric evaluates whether objects mentioned in generated captions are actually present in the image. Using ground-truth segmentations and reference captions from the MSCOCO dataset, CHAIR measures hallucination at two levels:
\begin{align}
\text{CHAIR}_{i} &= \frac{|\{\text{hallucinated objects}\}|}{|\{\text{all objects mentioned}\}|}, \\
\text{CHAIR}_{s} &= \frac{|\{\text{sentences with hallucinated objects}\}|}{|\{\text{all sentences}\}|}.
\end{align}
Here, $\text{CHAIR}_{i}$ quantifies hallucination at the object-instance level, while $\text{CHAIR}_{s}$ captures the proportion of captions containing hallucinated objects. Lower scores indicate better grounding and fewer hallucinated mentions.

We compute CHAIR scores on 500 randomly selected images from the MSCOCO validation set, using model-generated captions and official COCO object annotations as references.

\section{POPE: Probing Object-Presence Hallucination}
\label{appendix:hallucination_metrics:pope}

The \textit{POPE} benchmark probes whether models falsely predict the presence of objects not found in the image. Unlike CHAIR, which evaluates free-form captioning, POPE uses templated binary questions of the form: ``\textit{Is there a \{object\} in the image?}'' and compares the model’s answer to COCO annotations. The focus is on identifying \textit{false positives}, i.e., cases where the model incorrectly affirms the presence of objects that are absent.

POPE defines three targeted subsets to assess different sources of hallucination:

\begin{itemize}
    \item \textbf{Random}: Object categories are randomly sampled and paired with images in which the object is absent, probing baseline hallucination without context.
    
    \item \textbf{Adversarial}: Object prompts are selected to be semantically plausible within the scene (e.g., ``fork'' in a kitchen) but are not present, testing model susceptibility to co-occurrence priors.
    
    \item \textbf{Popular}: High-frequency object categories are paired with unrelated images to measure overprediction due to training frequency bias.
\end{itemize}

To ensure comparability with CHAIR, we evaluate POPE using questions corresponding to the same 500 randomly selected images from the MSCOCO validation set. For each model response, we compute standard classification metrics: \textbf{precision}, \textbf{recall}, \textbf{F1 score}, and \textbf{accuracy}, defined as follows:

\begin{align} \text{Precision} &= \frac{\text{TP}}{\text{TP} + \text{FP}}, &\text{Recall} &= \frac{\text{TP}}{\text{TP} + \text{FN}}, \\ \text{F1 Score} &= \frac{2 \cdot \text{Precision} \cdot \text{Recall}}{\text{Precision} + \text{Recall}}, &\text{Accuracy} &= \frac{\text{TP} + \text{TN}}{\text{TP} + \text{FP} + \text{TN} + \text{FN}}, \end{align}

where TP, FP, TN, and FN denote true/false positives and negatives. These metrics together capture hallucination behavior, balancing precision (avoiding false affirmations) with recall and accuracy (correctly rejecting absent objects).

% \section{Implementation Details for Baseline Comparisons}

% To enable fair comparisons across models, we evaluate hallucination mitigation methods including VCD, OPERA, and SPARC on both LLaVA-1.5 and Qwen2.5-VL. While official implementations are available for LLaVA, no publicly available implementations exist for these methods on Qwen2.5-VL. Accordingly, we re-implemented each method for Qwen2.5-VL following the descriptions in the original papers.

% Below, we provide the specific hyperparameters and implementation choices used in our Qwen2.5-VL experiments to ensure reproducibility and transparency.

\section{Evaluation on POPE Benchmark}
\label{appendix:pope_results}

In addition to CHAIR, we evaluate our structured scaffolding method using the POPE benchmark, which probes object hallucination through structured binary questions. As described in Appendix~\ref{appendix:hallucination_metrics:pope}, POPE defines three subsets—\textit{Random}, \textit{Adversarial}, and \textit{Popular}—each targeting different hallucination biases such as language priors, scene co-occurrence, and object frequency.

While our proposed strategy is designed to strengthen grounding during \textit{long-form generation}, we also evaluate on POPE to ensure coverage across widely used hallucination benchmarks. Although POPE is less aligned with our method’s strengths, results remain robust and comparable, demonstrating that the approach generalizes beyond descriptive settings.

We use the same 500 MSCOCO validation images as in the CHAIR evaluation. From the POPE benchmark, we select yes/no question subsets corresponding to these images. Each image has six questions per subset (Random, Adversarial, Popular), yielding \textbf{3,000 questions per subset} and \textbf{9,000 in total} per model. We evaluate both baseline and structured versions of LLaVA-1.5 and Qwen2.5-VL using standard binary classification metrics: precision, recall, F1 score, and accuracy.

\begin{table}[ht!]
\centering
\caption{Results of POPE evaluation on the MSCOCO dataset across Random, Popular, and Adversarial splits. Metrics reported are Accuracy, Precision, Recall, F1 Score, and Yes (\%) for LLaVA-1.5.}
\label{tab:pope_results_llava}
\resizebox{0.75\textwidth}{!}{
\begin{tabular}{llcccccc} 
\toprule
\textbf{Model} & \textbf{POPE} & \textbf{Method} & \textbf{Accuracy} & \textbf{Precision} & \textbf{Recall} & \textbf{F1 Score} & \textbf{Yes (\%)} \\
\midrule

\multirow{15}{*}{LLaVA-1.5} 
& \multirow{5}{*}{\textit{Random}}
  & Base     & \underline{87.27} & \underline{97.47} & \underline{76.66} & \underline{85.82} & 39.53 \\
& & OPERA    & 86.17 & 97.00 & 74.88 & 84.52 & 38.93 \\
& & VCD      & 85.43 & 93.86 & 75.99 & 83.99 & 40.70 \\
& & SPARC    & 85.10 & \textbf{98.90} & 71.34 & 82.89 & 36.50 \\
& & \textbf{Structured}     & \textbf{89.23} & 95.19 & \textbf{82.75} & \textbf{88.53} & 43.67 \\
\cmidrule(l){2-8} 

& \multirow{5}{*}{\textit{Popular}}
  & Base     & \underline{85.93} & \underline{94.28} & \underline{76.63} & \underline{84.54} & 40.80 \\
& & OPERA    & 84.93 & 93.99 & 74.78 & 83.30 & 39.97 \\
& & VCD      & 83.67 & 89.89 & 76.06 & 82.40 & 42.53 \\
& & SPARC    & 83.67 & \textbf{97.06} & 71.74 & 82.50 & 39.67 \\
& & \textbf{Structured}     & \textbf{86.07} & 88.77 & \textbf{82.77} & \textbf{85.67} & 46.90 \\
\cmidrule(l){2-8}

& \multirow{5}{*}{\textit{Adversarial}}
  & Base     & 81.43 & 86.17 & 75.15 & 80.28 & 43.87 \\
& & OPERA    & 82.07 & 87.76 & 74.88 & \underline{80.81} & 43.03 \\
& & VCD      & 80.57 & 83.94 & \underline{76.01} & 79.78 & 45.67 \\
& & SPARC    & \textbf{84.93} & \textbf{98.90} & 70.98 & 82.64 & 36.27 \\
& & \textbf{Structured}    & \underline{83.50} & \underline{89.03} & \textbf{76.72} & \textbf{82.42} & 43.43 \\
\bottomrule
\end{tabular}
\textbf{}}
\end{table}

\begin{table}[ht!]
\centering
\caption{Results of POPE evaluation on the MSCOCO dataset across Random, Popular, and Adversarial splits. Metrics reported are Accuracy, Precision, Recall, F1 Score, and Yes (\%) for Qwen2.5-VL.}
\label{tab:pope_results_qwen}
\resizebox{0.75\textwidth}{!}{
\begin{tabular}{llcccccc} % <-- 8 columns: ll + six c's
\toprule
\textbf{Model} & \textbf{POPE} & \textbf{Method} & \textbf{Accuracy} & \textbf{Precision} & \textbf{Recall} & \textbf{F1 Score} & \textbf{Yes (\%)} \\
\midrule

\multirow{15}{*}{Qwen2.5-VL} % <-- 3 blocks * 5 rows = 15
& \multirow{5}{*}{\textit{Random}}
  & Base     & \underline{89.73} & 98.39 & \underline{80.91} & \underline{88.80} & 41.37 \\
& & OPERA    & 88.83 & \textbf{98.76} & 78.85 & 87.69 & 40.27 \\
& & VCD      & \textbf{89.90} & \underline{98.55} & \textbf{81.13} & \textbf{88.99} & 41.43 \\
& & SPARC    & 85.33 & 95.91 & 71.76 & 82.10 & 35.07 \\
& & \textbf{Structured}     & 85.67 & 98.22 & 72.90 & 83.69 & 37.43 \\
\cmidrule(l){2-8} 

& \multirow{5}{*}{\textit{Popular}}
  & Base     & \underline{88.83} & 96.07 & \underline{81.09} & \underline{87.95} & 42.40 \\
& & OPERA    & 87.40 & 96.70 & 77.63 & 86.12 & 40.43 \\
& & VCD      & \textbf{89.00} & 96.09 & \textbf{81.44} & \textbf{88.16} & 42.63 \\
& & SPARC    & 84.77 & \underline{96.88} & 72.02 & 82.62 & 37.37 \\
& & \textbf{Structured}     & 85.07 & \textbf{96.92} & 72.72 & 83.09 & 37.87 \\
\cmidrule(l){2-8}

& \multirow{5}{*}{\textit{Adversarial}}
  & Base     & \textbf{87.43} & 93.16 & 81.02 & \underline{86.66} & 43.83 \\
& & OPERA    & 86.43 & \textbf{94.35} & 77.62 & 85.17 & 41.30 \\
& & VCD      & \underline{87.40} & 92.64 & \underline{81.49} & \textbf{86.71} & 44.37 \\
& & SPARC    & 81.17 & 80.35 & \textbf{82.75} & 81.53 & 51.73 \\
& & \textbf{Structured}     & 83.73 & \underline{93.60} & 72.65 & 81.80 & 39.07 \\
\bottomrule
\end{tabular}
}
\end{table}

Importantly, we observe that the structured version achieves lower \textbf{recall} than the baseline. Analyzing the false negatives reveals systematic patterns, such as objects positioned across grid boundaries or large objects spanning multiple cells. In this work we adopt a general prompting strategy, but more tailored prompts that explicitly define such corner cases could further reduce false negatives. Additionally, techniques such as ensembling over diverse structural partitions may help compensate for the limitations of any single structure.

The reduced recall on POPE is also consistent with the task design. Unlike long-form captioning, POPE is a discriminative benchmark where the model answers a single yes/no question about one queried object. In our analysis, external cues are most effective in settings that require sustained grounding across many objects and long responses. This trend is reflected in Table~\ref{tab:synthetic_hallucination}, where the gains increase as the number of objects grows. Similarly, the cross-attention analysis on MS-COCO in Fig.~\ref{fig:attention_span_coco} shows that structured inputs mainly improve grounding in later decoding stages, where baseline attention to the image has already decayed.

We also evaluate a multi-scale partitioning strategy motivated by the reviewer suggestion. Instead of using a single fixed structure, we ensemble predictions from five different partition layouts: $2{\times}2$, $3{\times}3$, and $4{\times}4$ grids, together with 3-row and 4-row horizontal partitions. Final predictions are obtained using majority voting across structures. Table~\ref{tab:pope_ensemble} shows that this strategy consistently improves performance over the single-structure setup, particularly in recall, suggesting that adaptive or multi-scale cues help reduce failures caused by object-boundary ambiguity.

\begin{table}[ht]
\centering
\caption{POPE results for Qwen2.5-VL using multi-scale structured cues with majority voting.}
\label{tab:pope_ensemble}
\resizebox{0.65\linewidth}{!}{
\begin{tabular}{llccccc}
\toprule
\textbf{Method} & \textbf{Subset} & \textbf{Accuracy} & \textbf{Precision} & \textbf{Recall} & \textbf{F1} & \textbf{Yes(\%)}\\
\midrule
\multirow{3}{*}{Ensemble}
& Adversarial & 85.72 & 93.96 & 76.59 & 84.40 & 41.08\\
& Random      & 87.82 & 99.06 & 76.48 & 86.31 & 38.78\\
& Popular     & 86.82 & 96.41 & 76.66 & 85.41 & 40.00\\
\midrule
\multirow{3}{*}{Structured}
& Adversarial & 83.73 & 93.60 & 72.65 & 81.80 & 39.08\\
& Random      & 85.67 & 98.22 & 72.90 & 83.69 & 37.42\\
& Popular     & 85.07 & 96.92 & 72.72 & 83.09 & 37.86\\
\bottomrule
\end{tabular}
}
\end{table}

\section{Evaluation on ROPE Benchmark}
\label{appendix:rope_results}

To further evaluate hallucination in multi-object settings, we conduct experiments on the ROPE benchmark \cite{rope}. Unlike POPE, which focuses on single-object yes/no queries, ROPE evaluates hallucination under multi-object conditions and includes several target objects per image.
ROPE contains three evaluation settings: \textit{Homogeneous}, \textit{Heterogeneous}, and \textit{Wild}. The benchmark provides bounding boxes for target objects, which already serve as spatial external cues. Following our structured prompting setup, we additionally assign a unique symbol to each bounding box and include the same symbols in the textual prompt to encourage aligned multimodal grounding.

Table~\ref{tab:rope_results} reports the results on the ROPE benchmark. The structured version consistently improves over the baseline across all settings, with the largest gain observed in the more challenging heterogeneous split. These results are consistent with our observations on the synthetic datasets, where external cues become more beneficial as the number of objects and grounding complexity increase.
While the proposed method improves performance, the gains remain moderate because ROPE already includes explicit spatial cues through bounding boxes. Nevertheless, the results suggest that aligned multimodal cues remain beneficial even in more realistic multi-object hallucination settings.

\begin{table}[ht]
\centering
\caption{Results on the ROPE benchmark \cite{rope}.}
\label{tab:rope_results}
\resizebox{0.45\linewidth}{!}{
\begin{tabular}{lccc}
\toprule
\textbf{Method} & \textbf{Wild} & \textbf{Hom.} & \textbf{Het.} \\
\midrule
Baseline & 63.13 & 76.10 & 43.49 \\
Grounding IDs & \textbf{66.63} & \textbf{78.60} & \textbf{49.88} \\
\bottomrule
\end{tabular}
}
\end{table}

\section{Hyperparameters for Hallucination-Mitigation Methods}
\label{appendix:hyperparams_hallucination}

For all hallucination experiments on MSCOCO (CHAIR and POPE), we use a unified decoding configuration unless otherwise specified: greedy decoding with temperature $T=0.0$, top-$p=1.0$, beam size $1$, and a maximum of $512$ new tokens.

Below we list the exact hyperparameters used for each hallucination-mitigation baseline, following the recommendations of the original papers.

\paragraph{VCD (Visual Contrastive Decoding).}
We follow the hyperparameters suggested in the VCD paper. The decoding coefficients are fixed to:
\[
\alpha = 1,\qquad \beta = 0.1,\qquad \gamma = 0.1.
\]
For the diffusion-based distortion process, we adopt the recommended number of noise steps, $
T = 500.$
These settings are used for both LLaVA-1.5 and Qwen2.5-VL.

\paragraph{OPERA.}
We use the default configuration recommended in the OPERA paper. The method introduces an over-trust logit penalty and a retrospection-allocation mechanism, with parameters:
\[
N_{\text{can}} = 5, \qquad \sigma = 50, \qquad \alpha = 1, \qquad r = 15.
\]
These unified settings are applied across all evaluated models, including LLaVA-1.5 and Qwen2.5-VL. We use beam size $5$ following the authors' implementation.

\paragraph{SPARC.}
For SPARC, we follow the recommended hyperparameters from the original paper. Global parameters include:
\[
\alpha = 1.1,\qquad \beta = 0.1.
\]
Attention recalibration is applied across all layers. The token-selection threshold $\tau$ and the attention-extraction layer depend on the model:
\begin{itemize}
    \item \textbf{LLaVA-1.5:} $\tau = 1.5$, selected layer $= 20$.
    \item \textbf{Qwen2.5-VL:} $\tau = 3.0$, selected layer $= 18$.
\end{itemize}

\FloatBarrier
\section{Activation Swapping}
\label{app:act-swap}

\paragraph{Data generation.}
We generate 100 synthetic scenes (“samples”). Each scene is a four-row grid with exactly one \emph{symbol} and one \emph{object} per row. We draw four \emph{distinct} shapes (e.g., circle, diamond, square, triangle) and eight \emph{distinct} colors (chosen without replacement from a nine-color palette) and assign one unique (shape, color) pair to each row. Symbols are randomly permuted over the physical rows (top$\rightarrow$bottom) in every sample.

\paragraph{Pairing and target selection.}
For each sample $x$, we select a paired sample $x'$ and two distinct target symbols. Because each scene contains four unique (shape, color) pairs, targets are unambiguous within a scene. To isolate intervention effects, we impose a \emph{symmetric mismatch} between $x$ and $x'$: for each chosen symbol $s$, the associated object in $x$ and the associated object in $x'$ must differ in \emph{both} shape and color (i.e., $\text{shape}_x(s)\neq\text{shape}_{x'}(s)$ and $\text{color}_x(s)\neq\text{color}_{x'}(s)$). This design also rules out the \emph{symbol-activation} alternative hypothesis—namely, that the model might answer by reading object attributes cached in the symbol’s own representation. Because our swaps patch \textbf{only object image tokens} with a one-token dilation (\textbf{pad}~=~1) and \emph{never} include symbol tokens, and because paired objects differ on both attributes, any post-swap answer that follows the injected attributes cannot be explained by information stored in symbol activations.

\paragraph{Intervention.}
For each pair $(x,x')$: (i) we first query both contexts \emph{without intervention} to obtain baseline answers of the form “What is the \emph{shape} / \emph{color} of the object associated with symbol $S$?”; (ii) we then perform an \emph{object-only activation swap}. Let $\mathcal{I}_r$ be the set of \emph{image-token} indices belonging to the \textbf{object} in the referenced row $r$ (obtained by mapping patch centers to the object’s bounding box); we dilate this mask by one token in each axis to get $\tilde{\mathcal{I}}_r$ (\textbf{pad}=1). We swap the hidden activations at $\tilde{\mathcal{I}}_r$ between the two runs and leave all other positions \emph{unchanged}—in particular, we do \emph{not} alter (a) any \textbf{text tokens} that mention symbols, or (b) image tokens occupied by the \textbf{symbol glyphs} themselves. (iii) We re-ask the same questions and record post-swap answers for the two targets.

\paragraph{Outputs.}
For each context we log pre-/post-swap answers and token-level logits for the queried attributes.

\paragraph{Prompt.}
For each target symbol, we use the same wording across both contexts and substitute \texttt{<row\_symbol>} with one of \texttt{\&}, \texttt{\$}, \texttt{\#}, \texttt{@}:
\begin{quote}\small\ttfamily
Scan the image using the symbols on the left (\&, \$, \#, @) as row labels.\\
What is the shape of the object in the ``<row\_symbol>'' row?
\end{quote}
(For color queries, replace “shape” with “color”.)

% ----------------------------------------------------------------------

\section{Disjoint-Symbol Swap}
\label{sec:disjoint_swap}
To further assess the role of abstract identifiers, we repeat the activation-swap experiment with \emph{disjoint symbol sets} across contexts. The source image uses \{\&,\$,\#,@\}, while the target image uses \{+, $\times$, \%, !\}, with no overlap (Fig. \ref{fig:disjoint_swap_stimuli}). After object-only activation patching from the source into the target (pad $=1$; symbol tokens never swapped), we \emph{intentionally query the target} using a \textbf{source} symbol (e.g., \&), which is not physically present in the target image. The query omits any enumeration of the available symbols so as not to contradict the target’s visual glyphs.

\paragraph{Prompt.}
We use the same wording as above, but without listing the four symbols. For a chosen \texttt{<row\_symbol>} (e.g., \texttt{\&}) we ask:
\begin{quote}\small\ttfamily
Scan the image using the symbols on the left as row labels.\\
What is the shape of the object in the ``<row\_symbol>'' row?\\
If there is no object corresponding to ``<row\_symbol>'' in the image, answer \emph{none}.
\end{quote}
(For color queries, replace “shape” with “color”.)

\paragraph{Bias control and evaluation protocol.}
Allowing the response \emph{none} mitigates bias toward hallucinating an answer when the queried symbol is not present. For evaluation, the accuracy reported in the paper is computed on the subset of examples where, in the \emph{baseline} (non-intervened) condition, the model correctly answered \emph{none} for the queried symbol. This isolates the effect of the intervention from cases where the model would have produced a spurious non-none answer even without patching.

% \Needspace*{14\baselineskip}
% \vspace{-0.2\baselineskip}
% \begin{figure}[H]
%   \captionsetup{skip=3pt} % tighter caption spacing
%   \centering
%   \begin{subfigure}{0.42\linewidth}
%     \centering
%     \includegraphics[width=\linewidth]{figs_appendix/tgt_symbols_disjoint_f.pdf}
%     \caption{Source: symbols \&,\$,\#,@.}
%   \end{subfigure} \hspace{0.02\textwidth}
%   \begin{subfigure}{0.42\linewidth}
%     \centering
%     \includegraphics[width=\linewidth]{figs_appendix/src_symbols_disjoint_f.pdf}
%     \caption{Target: symbols !,\%,$\times$,+.}
%   \end{subfigure}
%   \caption{Disjoint-symbol control for the activation-swap test. Object-image activations are transferred with pad $=1$; symbol tokens are never swapped. The target is queried with a \emph{source} symbol that does not appear in the image, and the prompt allows the answer \emph{none} to avoid bias.}
%   \label{fig:disjoint_swap_stimuli}
% \end{figure}

\section{Extended Intervention Analysis Beyond the Controlled Setting}
\label{app:extended_intervention}

The activation-swapping experiments in Section~\ref{sec:causality} use a controlled setting with one object per row to isolate object-symbol binding effects. Here, we extend the analysis to more complex synthetic layouts and natural images.

\begin{figure}[H]
    \centering
    \includegraphics[width=0.62\linewidth]{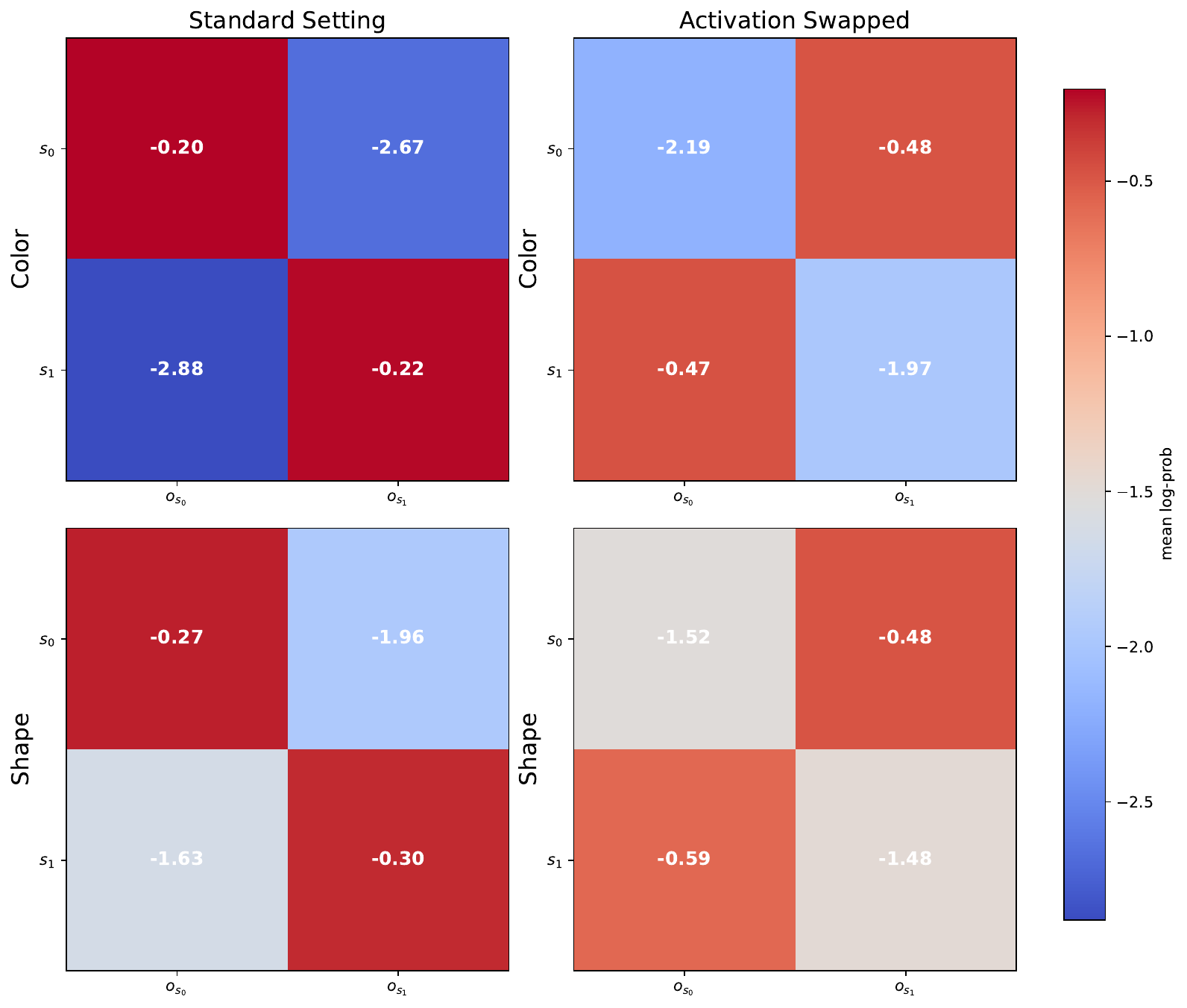}
    \vspace{-2.5mm}
    \caption{
    Average log probabilities in the multi-object-per-row activation-swapping setting, reported separately for color and shape. In the standard setting, the highest probabilities align with the original symbol--object associations. After activation swapping, the probability mass shifts toward the transferred associations for both attributes.
    }
    \label{fig:multi_object_swap_shape_color}
    \vspace{-1mm}
\end{figure}

\paragraph{Multiple objects per partition.}
We first evaluate a harder synthetic setting where each row contains multiple objects instead of a single object. The same activation-swapping procedure from Section~\ref{sec:causality} is applied. After swapping object activations between contexts, the model continues to follow the transferred symbol-object bindings rather than local spatial placement.
Using the same evaluation protocol, swap accuracy remains high at $\mathbf{0.83}$, compared to $\mathbf{0.98}$ in the single-object-per-row setup. Figure~\ref{fig:multi_object_swap_shape_color} shows the average log probabilities separately for color and shape before and after intervention. In the standard setting, the model assigns higher probability to the original symbol--object associations, while after activation swapping the high-probability entries shift toward the transferred bindings for both attributes.

\paragraph{Natural-image intervention analysis.}
Direct activation swapping on natural images is difficult because transferring visual tokens across scenes introduces confounding factors such as object scale, background variation, and clutter. Instead, we use an alternative intervention analysis based on PatchScopes~\cite{patchscope}.

We apply the same structured cues to MS-COCO images and use annotated bounding boxes to extract visual tokens corresponding to individual objects. The object tokens are then transferred into an otherwise empty visual context, and the model is queried using the corresponding symbol.
If Grounding IDs are encoded in the transferred representations, the model should recover the symbol-object association from the transferred tokens alone. Using this procedure, the model achieves a decoding accuracy of $\mathbf{0.76}$, substantially above the random baseline of $0.25$. Although weaker than the synthetic setting, this result suggests that the learned bindings persist in more realistic scenes beyond the controlled row-based layouts.

\section{Extended Experiments on Alternative Models}\label{sec:extended_experiments}

\begin{figure}[htb]
    \centering

    \begin{subfigure}{0.28\textwidth}
        \includegraphics[width=\linewidth]{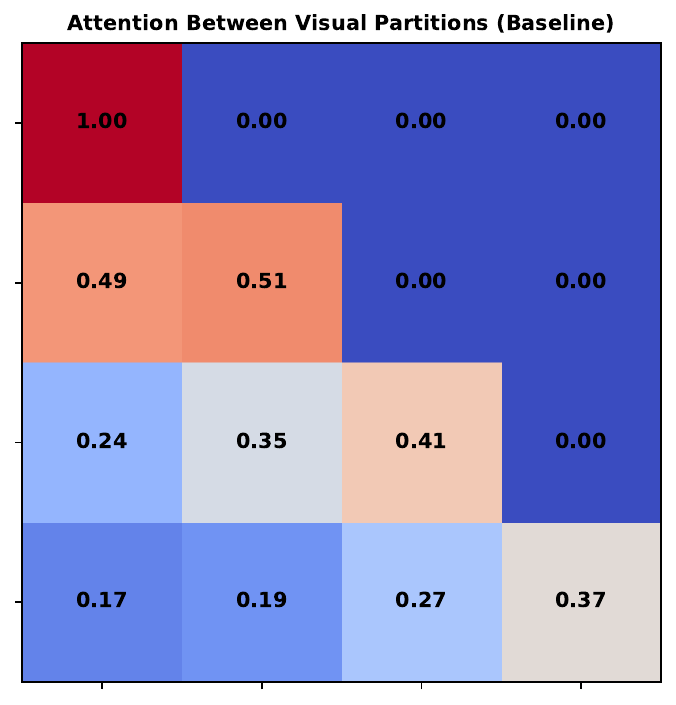}
        \vspace{-2.5mm}
    \end{subfigure}
    \hspace{0.02\textwidth}
    \begin{subfigure}{0.28\textwidth}
        \includegraphics[width=\linewidth]{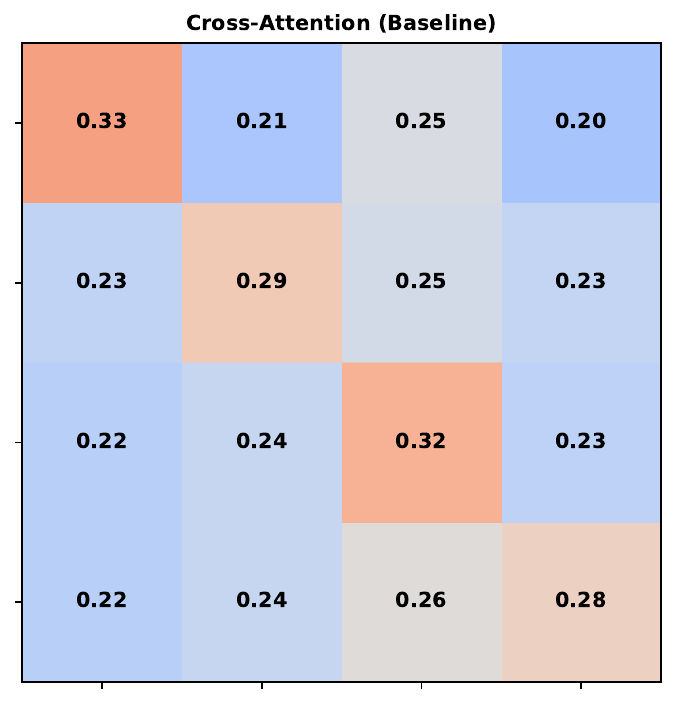}
        \vspace{-2.5mm}
    \end{subfigure}

    \vspace{0.1cm}

    \begin{subfigure}{0.28\textwidth}
        \includegraphics[width=\linewidth]{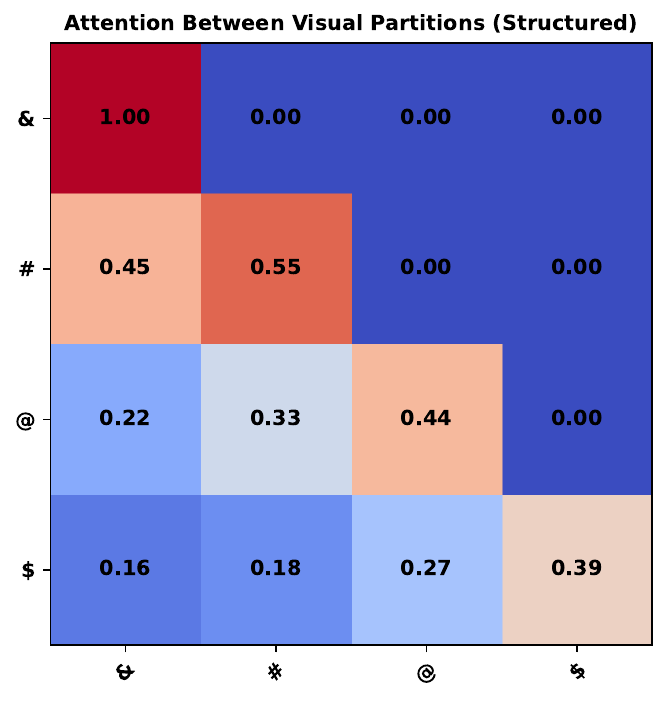}
        \vspace{-3mm}
        \caption{}
    \end{subfigure}
    \hspace{0.02\textwidth}
    \begin{subfigure}{0.28\textwidth}
        \includegraphics[width=\linewidth]{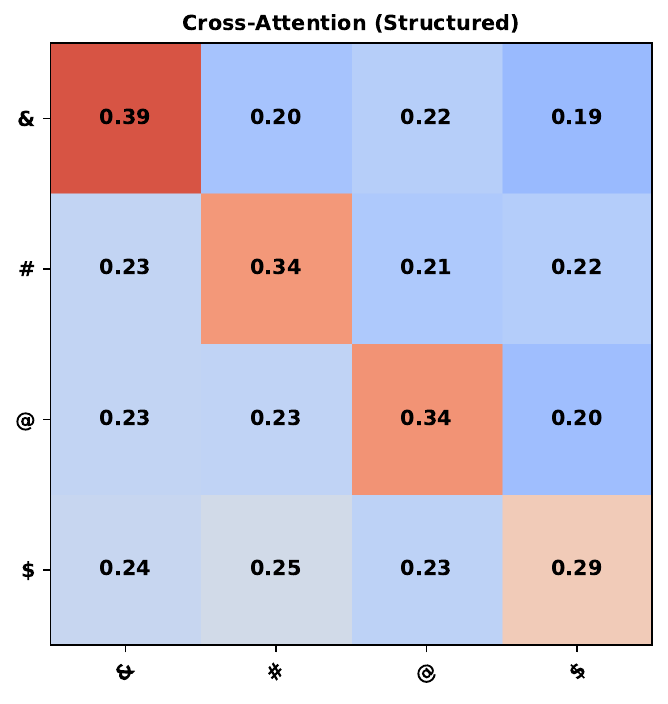}
        \vspace{-3mm}
        \caption{}
    \end{subfigure}

    \caption{
    Attention patterns of InternVL3.5 (8B) under baseline (top) and structured (bottom) inputs for the scene description task.  
    (a) Within-modality visual attention.  
    (b) Cross-modality attention.
    }
    \label{fig:attention_mat_internvl}
\end{figure}

\begin{figure}[htb]
    \centering

    \begin{subfigure}{0.28\textwidth}
        \includegraphics[width=\linewidth]{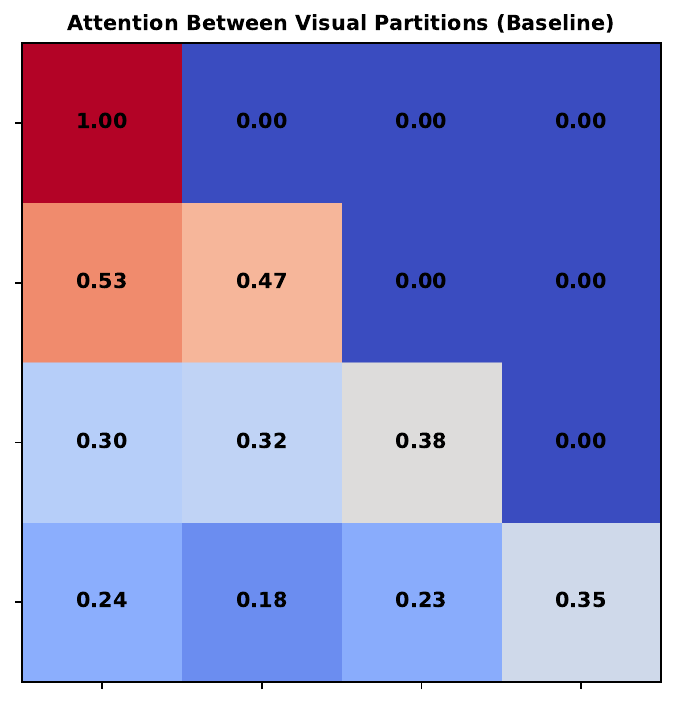}
        \vspace{-2.5mm}
    \end{subfigure}
    \hspace{0.02\textwidth}
    \begin{subfigure}{0.28\textwidth}
        \includegraphics[width=\linewidth]{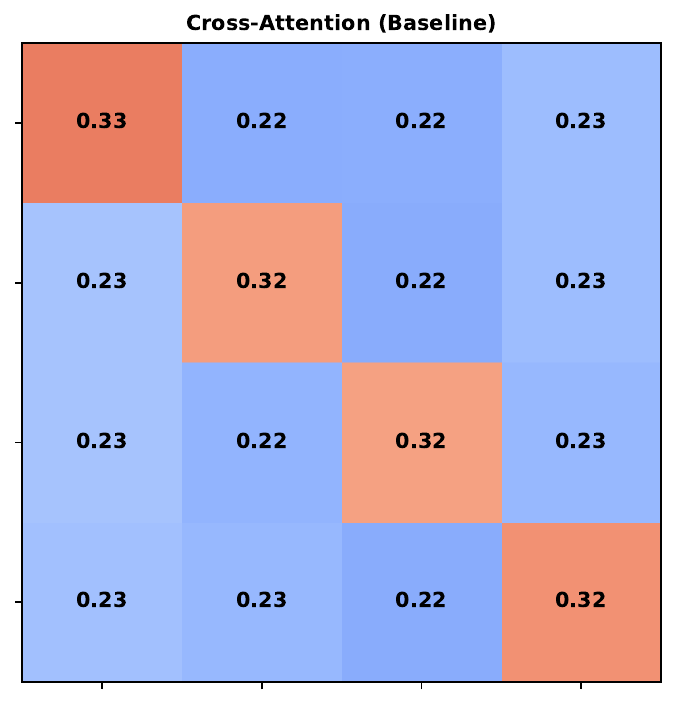}
        \vspace{-2.5mm}
    \end{subfigure}

    \vspace{0.1cm}

    \begin{subfigure}{0.28\textwidth}
        \includegraphics[width=\linewidth]{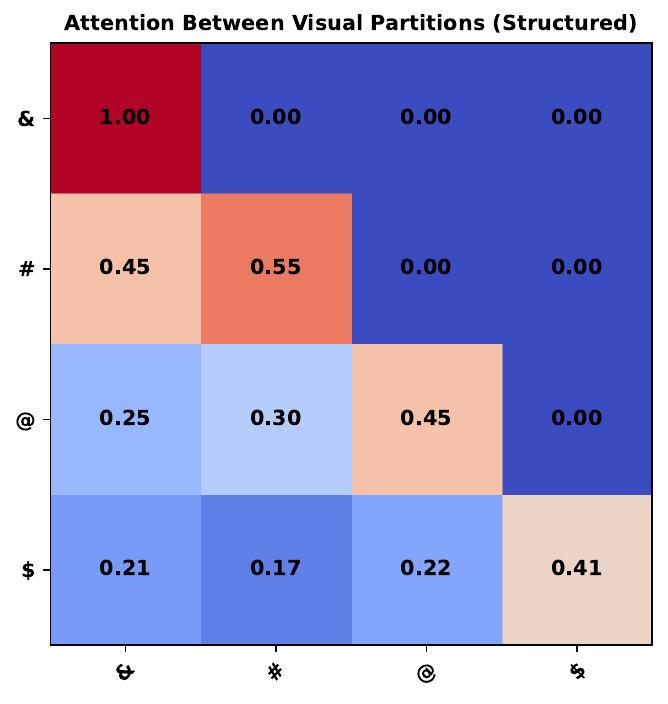}
        \vspace{-3mm}
        \caption{}
    \end{subfigure}
    \hspace{0.02\textwidth}
    \begin{subfigure}{0.28\textwidth}
        \includegraphics[width=\linewidth]{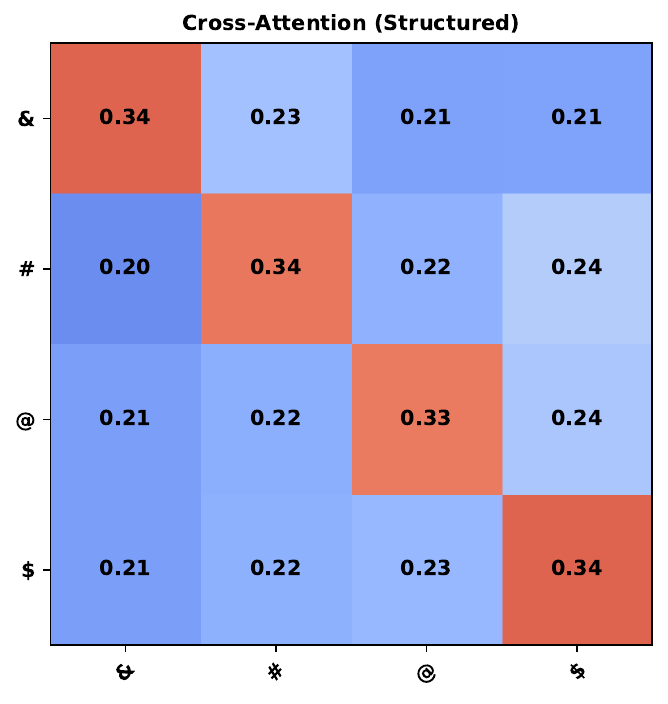}
        \vspace{-3mm}
        \caption{}
    \end{subfigure}

    \caption{
    Attention patterns of Qwen2.5-VL (3B) under baseline (top) and structured (bottom) inputs for the scene description task.  
    (a) Within-modality visual attention.  
    (b) Cross-modality attention.
    }
    \label{fig:attention_mat_qwen3b}
\end{figure}

Experiments in Sections~\ref{sec:existence} and~\ref{sec:causality} were originally conducted on Qwen2.5-VL (7B). Here, we present additional experiments on models with different sizes and architectures. Figures~\ref{fig:attention_mat_internvl} and~\ref{fig:attention_mat_qwen3b} show within- and cross-modality attention analyses for InternVL3.5 (8B) and Qwen2.5-VL (3B), which exhibit the same attention progression as Qwen2.5-VL (7B) (Fig.~\ref{fig:attention_mat}).

\begin{figure}[t]
    \centering

    \begin{subfigure}{0.49\textwidth}
        \includegraphics[width=\linewidth]{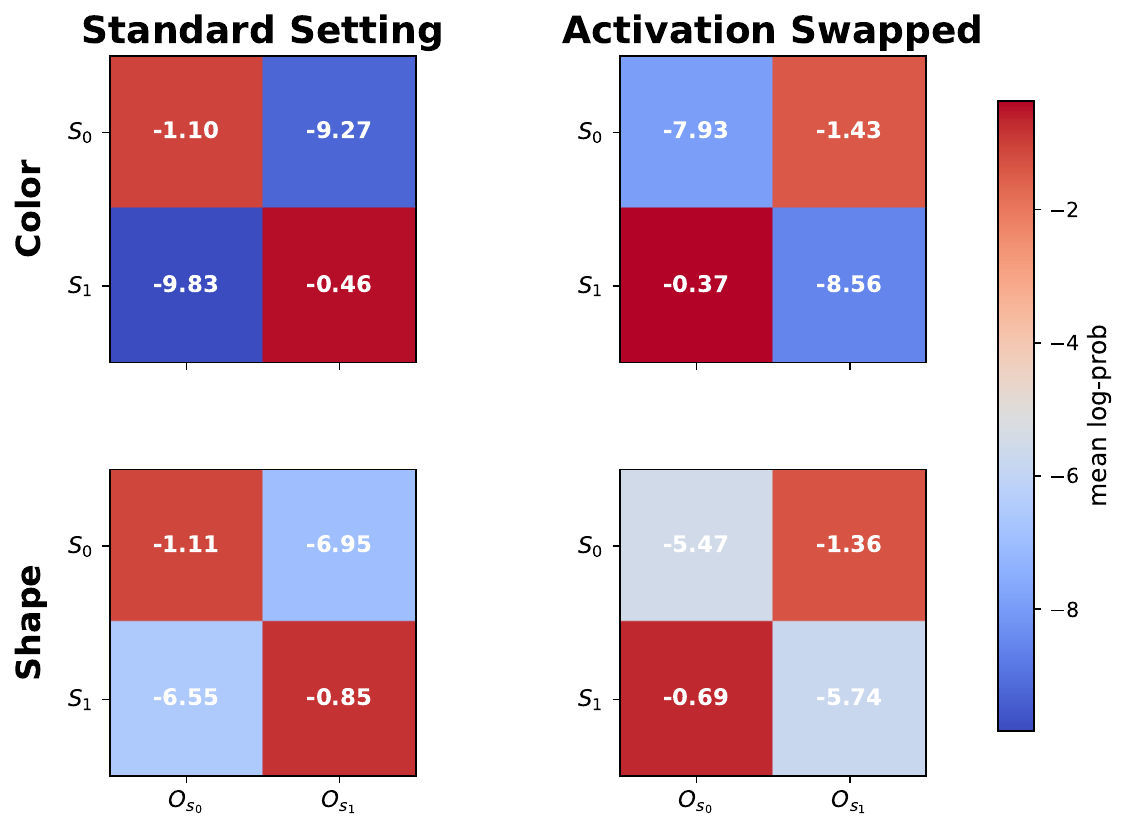}
        \vspace{-2.5mm}
        \caption{}
    \end{subfigure}
    \begin{subfigure}{0.49\textwidth}
        \includegraphics[width=\linewidth]{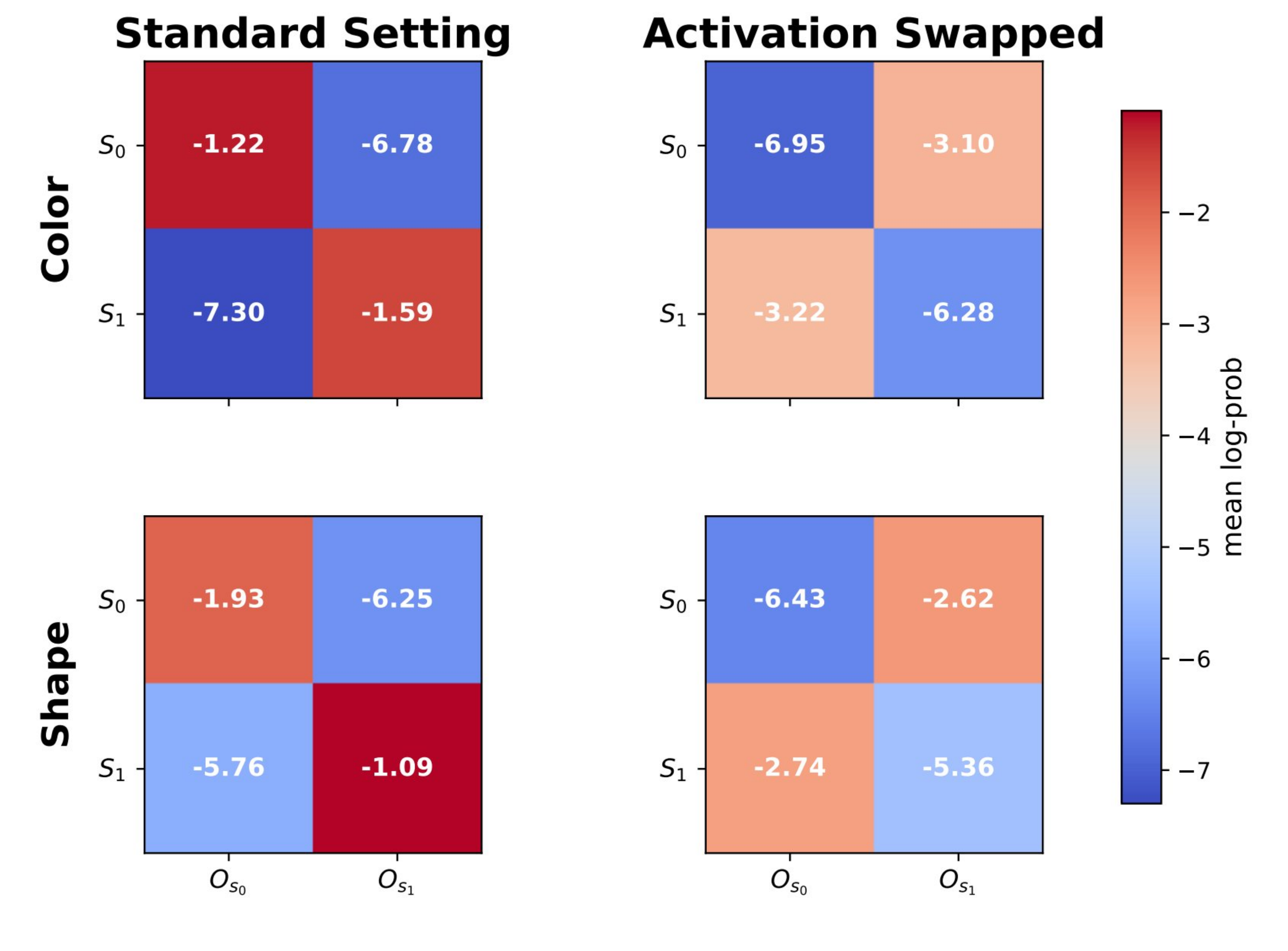}
        \vspace{-2.5mm}
        \caption{}
    \end{subfigure}
    \caption{
    Average log probabilities of $c$ and $c^*$ over valid row–symbol–object combinations for InternVL3.5 (a) and Qwen2.5-VL 3B (b). Rows and columns correspond to the selected query symbols and their associated objects.
    }
    \label{fig:swap_results_internvl_qwen3b}
\end{figure}

We also repeated the activation-swap experiment from Section~4 on InternVL3.5 (8B) and Qwen2.5-VL (3B). The models exhibit the same qualitative behavior as Qwen2.5-VL (7B). For InternVL3.5, swap accuracy is 0.75 for shape and 0.82 for color, well above the random chance level of 0.25 for shape and 0.125 for color. For Qwen2.5-VL (3B), swap accuracy is 0.93 for shape and 0.73 for color, indicating a consistent dependence on Grounding IDs. Fig.~\ref{fig:swap_results_internvl_qwen3b} shows the corresponding log-prob diagrams, which follow the same pattern observed in Qwen2.5-VL (7B) (Fig.~\ref{fig:swap_result}).

Finally, we repeat the hallucination mitigation experiments on MS-COCO using two additional open-source models of different sizes: Qwen2.5-VL (3B) and InternVL3.5 (8B). Results are reported in Table~\ref{tab:chair_internvl_qwen}.

\clearpage
\begin{table}[t]
\centering
\begin{tabular}{l l c c}
\toprule
Model & Method & CHAIR$_s\downarrow$ & CHAIR$_i\downarrow$ \\
\midrule
\multirow{2}{*}{Qwen2.5-VL 3B}
  & Baseline  & \underline{32.40} & \underline{8.82} \\
  & \textbf{Structured}   & \textbf{24.40} & \textbf{7.32} \\
\midrule
\multirow{2}{*}{InternVL3.5 8B}
  & Baseline  & \underline{32.40} & \textbf{6.09} \\
  & \textbf{Structured}   & \textbf{26.40} & \underline{8.08} \\
\bottomrule
\end{tabular}

\caption{CHAIR results on 500 MS-COCO samples for Qwen2.5-VL (3B) and InternVL3.5 (8B).}
\label{tab:chair_internvl_qwen}
\end{table}

\section{Clarification of Terminology}
\label{app:terminology}

Several terms used in the main paper relate to mechanisms that have been discussed in prior work on internal representations in language and vision–language models. For clarity, we provide brief descriptions here.

\paragraph{Symbolic.}
We use the term \emph{symbolic} to refer to the fact that external cues (such as characters or markers) function as explicit and discrete symbols that index different regions of the input. They can be consistently referenced by the model both in text and in internal computation.

\paragraph{Identifier.}
An \emph{identifier} denotes an internal code that the model assigns to all tokens associated with the same cue. This identifier allows the model to bind visual and textual elements referring to the same partition.

\paragraph{Latent.}
The identifier is \emph{latent} because it does not appear explicitly in the visible input. While the external cue is visible, the identifier itself arises in the model’s hidden activations and can be detected through probing and representation analysis.

\paragraph{Vector.}
We refer to the identifier as a \emph{vector} because it is expressed as a direction or component in activation space rather than as a discrete token.

\paragraph{Abstract.}
The identifier is \emph{abstract} because it does not encode perceptual details such as color or shape. Instead, it represents the relational property of membership in a specific partition, independent of the visual content of that partition.

\end{document}